\documentclass[preprint,12pt]{elsarticle} 

\usepackage[utf8]{inputenc}

\usepackage{amsmath,amssymb}

\usepackage{graphicx}
\usepackage{subfig}
\usepackage{booktabs}
\usepackage{multirow}
\usepackage{makecell}
\usepackage[table,xcdraw]{xcolor}

\usepackage{algorithm}
\usepackage{algpseudocode}

\usepackage{enumitem}
\usepackage{xurl}
\usepackage{url}

\usepackage{hyperref}
\hypersetup{
  colorlinks=true,
  linkcolor=blue,
  citecolor=blue,
  urlcolor=blue
}

\journal{Neurocomputing}

\begin{document}

\begin{frontmatter}

\title{BotaCLIP: Contrastive Learning for Botany-Aware Representation of Earth Observation Data}

\author[leca]{Selene Cerna\corref{cor1}}
\cortext[cor1]{Corresponding author.}
\ead{selene.cerna@univ-grenoble-alpes.fr}

\author[leca]{Sara Si-Moussi}
\author[leca]{Wilfried Thuiller}
\author[inria]{Hadrien Hendrikx}
\author[leca]{Vincent Miele}

\affiliation[leca]{organization={Univ. Grenoble Alpes, Univ. Savoie Mont Blanc, CNRS, LECA},
                   city={Grenoble},
                   country={France}}

\affiliation[inria]{organization={Centre Inria de l’Université Grenoble Alpes, CNRS, LJK},
                    city={Grenoble},
                    country={France}}

\begin{abstract}



Plants sustain terrestrial life and are key indicators for biodiversity monitoring, yet ecological mapping is limited by the mismatch between sparse in-situ vegetation surveys (relev\'es) and dense but indirectly informative Earth Observation (EO) imagery. EO foundation models yield transferable embeddings, but they are optimized for generic EO objectives and adapting them to ecological applications often requires costly end-to-end retraining. In this paper, we address this challenge by introducing BotaCLIP, a multimodal transfer framework that uses relev\'es as weak supervision to steer frozen EO vision models (DOFA, FLAIR, and DINOv3) toward biodiversity-relevant semantics via contrastive alignment. BotaCLIP trains lightweight projection adapters atop frozen EO encoders and a tabular encoder for high-dimensional plant community data (VEGETA), yielding a joint embedding space rich in transferable representations. A geometry-preserving regularizer further anchors the adapted image space to the frozen backbone's neighborhood structure, preventing over-specialization during alignment.

Motivated by real-world biodiversity modeling, we assess transferability in the French Alps across three tasks spanning ecological levels and classification/regression: plant presence, butterfly occurrence, and soil trophic group abundance prediction. Our results show that BotaCLIP features consistently outperform those from frozen EO vision models and a supervised baseline, offering a scalable, domain-aware approach to representation learning. By injecting ecological knowledge into frozen models, BotaCLIP not only enhances predictive performance, but also provides a generalizable pathway for leveraging foundation models in environmental science—where labeled data are limited and domain expertise is paramount.

\end{abstract}



\end{frontmatter}




\section{Introduction}

Plants underpin terrestrial ecosystems, driving primary productivity and sustaining biodiversity across trophic levels~\citep{CavenderBares2020,cardinale2012_bioanalyst}. As integrators of soil properties, microclimate, and disturbance histories~\citep{Chauvier2021}, plant communities are critical indicators for monitoring ecosystem functioning and guiding conservation and climate adaptation strategies \citep{Walker2014,IbarraManriquez2022}. Yet, the data streams available for large-scale monitoring remain fundamentally mismatched: in-situ vegetation surveys (relevés) provide rich, expert-curated descriptions of plant community composition, but are expensive and spatially sparse. Conversely, Earth Observation (EO) imagery provides dense spatial coverage, but only indirectly encodes the ecological processes and species assemblages. This mismatch is a bottleneck for biodiversity-oriented EO applications where the target signals, community structure, species assemblages, and functional traits, are poorly captured by standard remote-sensing labels.

Recent advances in EO foundation models and self-supervised encoders have shown strong transferability across standard benchmarks, spurring the use of pretrained embeddings as drop-in features for downstream tasks~\citep{nowakowski2025_transfer_learning_eo_review}. However, these representations are typically optimized for generic EO objectives (e.g., land cover classification) and lack explicit alignment with botanical composition or community-level semantics. As a result, purely image-derived embeddings may fail to capture the domain-specific patterns critical for ecological modelling, especially in data-scarce regimes where ecological labels are limited or costly to collect~\citep{trantas2025_bioanalyst}. The core challenge lies in adapting frozen EO models to encode ecological knowledge without prohibitive computational costs or end-to-end retraining.

We propose a paradigm shift: treating in-situ vegetation surveys as a source of weak supervision to guide pretrained EO representations toward domain-relevant semantics. To this end, we introduce BotaCLIP, a multimodal transfer learning framework that aligns aerial orthophotos with vegetation relevés via contrastive learning. It leverages lightweight projection adapters on top of pretrained EO vision models (DOFA~\citep{dofa}, FLAIR~\citep{flair}, and DINOv3~\citep{dinov3}, hereafter called EO backbones). BotaCLIP injects domain knowledge through VEGETA, a tabular encoder that compacts high-dimensional plant community descriptors for alignment with imagery. Unlike prior EO alignment methods, which rely on proxy signals like text, coordinates, and metadata, our approach directly exploits in-situ community composition, yielding a joint embedding space where orthophotos and relev\'es are semantically aligned. 

We focus on using the image-side embeddings as features for downstream ecological models, which reflects the practical setting of predicting on unseen orthophotos. To evaluate transferability, we benchmark BotaCLIP in the French Alps across three prediction tasks spanning ecological levels and methodological settings (classification/regression): plant presence prediction (assessing botanical signal recovery from imagery), butterfly occurrence prediction (testing transfer via vegetation–insect dependencies), and soil trophic group abundance prediction (probing aboveground–belowground linkages). Crucially, we design our experiments to reflect real-world constraints: embeddings are consumed by simple, practitioner-friendly models (e.g., Random Forests), ensuring that performance gains stem from representation quality rather than task-specific architectural complexity. Therefore, our work makes three key contributions:

\begin{itemize}

	\item A modular multimodal framework for ecological transfer learning. BotaCLIP aligns EO imagery with vegetation surveys using contrastive learning and lightweight adapters with geometry-preserving regularization to enable domain injection without end-to-end retraining.

    \item A controlled study of EO backbone transferability. We systematically evaluate three pretrained EO vision backbones to quantify how backbone choice influences ecological task performance, providing actionable insights for model selection in biodiversity applications

    \item Evidence of cross-ecological transfer. BotaCLIP-adapted embeddings outperform over both raw backbone and supervised baseline embeddings across all three tasks, demonstrating transfer beyond the training pairs and highlighting the framework’s potential for data-efficient environmental monitoring and modelling.

\end{itemize}


\section{Related Work}

Self-supervised and foundation-model approaches have rapidly reshaped EO representation learning~\cite{tuia2025_aiadvanceeo,nowakowski2025_transfer_learning_eo_review}. By leveraging large-scale unlabeled imagery and sensor diversity, pretrained EO encoders provide transferable features for many tasks, often outperforming training-from-scratch pipelines~\citep{dofa,flair,dinov3}. This trend has encouraged “embedding-first” workflows where downstream models consume frozen pretrained representations. Yet, most EO pretraining objectives and benchmarks focus on EO-centric semantics (e.g., land cover, urban mapping, change detection) rather than ecological community composition. For biodiversity applications, the critical gap is that ecological targets (species assemblages, functional structure, trophic dependencies) are not directly encoded by generic EO objectives, and require domain-guided adaptation to become accessible from imagery.

Long before modern representation learning, ecologists developed ordination methods to relate species composition to environmental gradients. Canonical Correspondence Analysis (CCA), for instance, learns coupled projections that maximize the correspondence between a community matrix (species presence-absence or abundance) and explanatory variables (e.g., climate, soil, topography)~\citep{terBraak1986_cca}. From a modern perspective, CCA can be viewed as an early form of multimodal coupling, aligning heterogeneous data views (community descriptors vs environmental descriptors) to uncover  shared latent factors driving ecological variation. This historical parallel highlights a long-standing need in ecology for methods that couple heterogeneous data views, a need that modern contrastive learning can address at scale, and that BotaCLIP explicitly builds upon. 



Recent advances in contrastive learning provide a scalable, non-linear counterpart to this coupling principle~\citep{li2025_mm_align_survey}. Contrastive learning has become a standard mechanism to align heterogeneous modalities in a shared embedding space by pulling together representations of paired observations and pushing apart mismatched pairs. Unlike supervised pipelines that fit task-specific predictors from labeled targets~\citep{Cimoli2024,SiMoussi2025_ContinentalScaleHabitat}, contrastive alignment can leverage weakly paired modalities to learn transferable representations with minimal task-specific supervision. The CLIP paradigm popularized contrastive alignment for image--text pairs~\citep{clip_radford2021}, and subsequent work extended it to non-linguistic supervision such as image--tabular pairing, multi-sensor alignment, and EO metadata~\citep{du2024,dofa,copernicusfm}. In EO, auxiliary non-text signals (e.g., time, geolocation, sensor characteristics) can provide semantic context and strengthen transferability under distribution shift~\citep{bourcier2024learning}. For example, GeoCLIP~\citep{geoclip} and SatCLIP~\citep{satclip} align images with coordinates or location embeddings to capture broad biogeographic structure without dense labels, while RemoteCLIP~\citep{remoteclip} shows that weak image--text pairing can inject high-level semantics into EO representations.
A particularly promising direction combines contrastive alignment with parameter-efficient adaptation, where frozen unimodal encoders are augmented with lightweight adapters~\citep{frozen_unimodal_alignment_cvpr2025}. This approach is ideal for EO data, where backbones are often large and expensive to fine-tune end-to-end.

Closer to our setting, several works have used biodiversity signals to supervise image representations. Models such as BioCLIP~\citep{bioclip} and CLIBD~\citep{clibd} align natural photos with taxonomic labels or DNA barcodes. Location-based frameworks such as WildSAT~\citep{wildsat} align satellite images with species occurrence maps from citizen science platforms, while EcoWikiRS ~\citep{Zermatten2025_EcoWikiRS} routes species observations through Wikipedia habitat descriptions to supervise aerial image encoders. These works establish that domain-specific biological signals can effectively guide learning of more transferable representations.



BotaCLIP advances this line of work by using vegetation relevés, which are exhaustive field surveys of plant community composition, as direct contrastive supervision for frozen EO encoders. Unlike generic EO metadata (e.g., acquisition time, coordinates), individual occurrence records, or text-mediated signals, relevés encode ecological structure directly through species identities, relative abundances, and co-occurrence patterns, offering a richer and more principled supervision source for steering EO embeddings toward biodiversity-relevant semantics. BotaCLIP further introduces a geometry-preserving regularizer that complements contrastive alignment by encouraging the adapted image space to retain the neighborhood structure of the pretrained EO backbone, mitigating over-specialization while preserving the rich representations already learned by the EO vision encoders models. By combining lightweight adapters, weak community-level supervision, and structure-preserving regularization, BotaCLIP provides a scalable, domain-aware framework for biodiversity-oriented EO representation learning.

\section{BotaCLIP for Botany-Aware Representation} 
\label{sec:botaclip_framework}

\subsection{Overview of the framework}
\label{sec:overview}

\begin{figure}[!ht]
\centering
\includegraphics[width=0.98\linewidth]{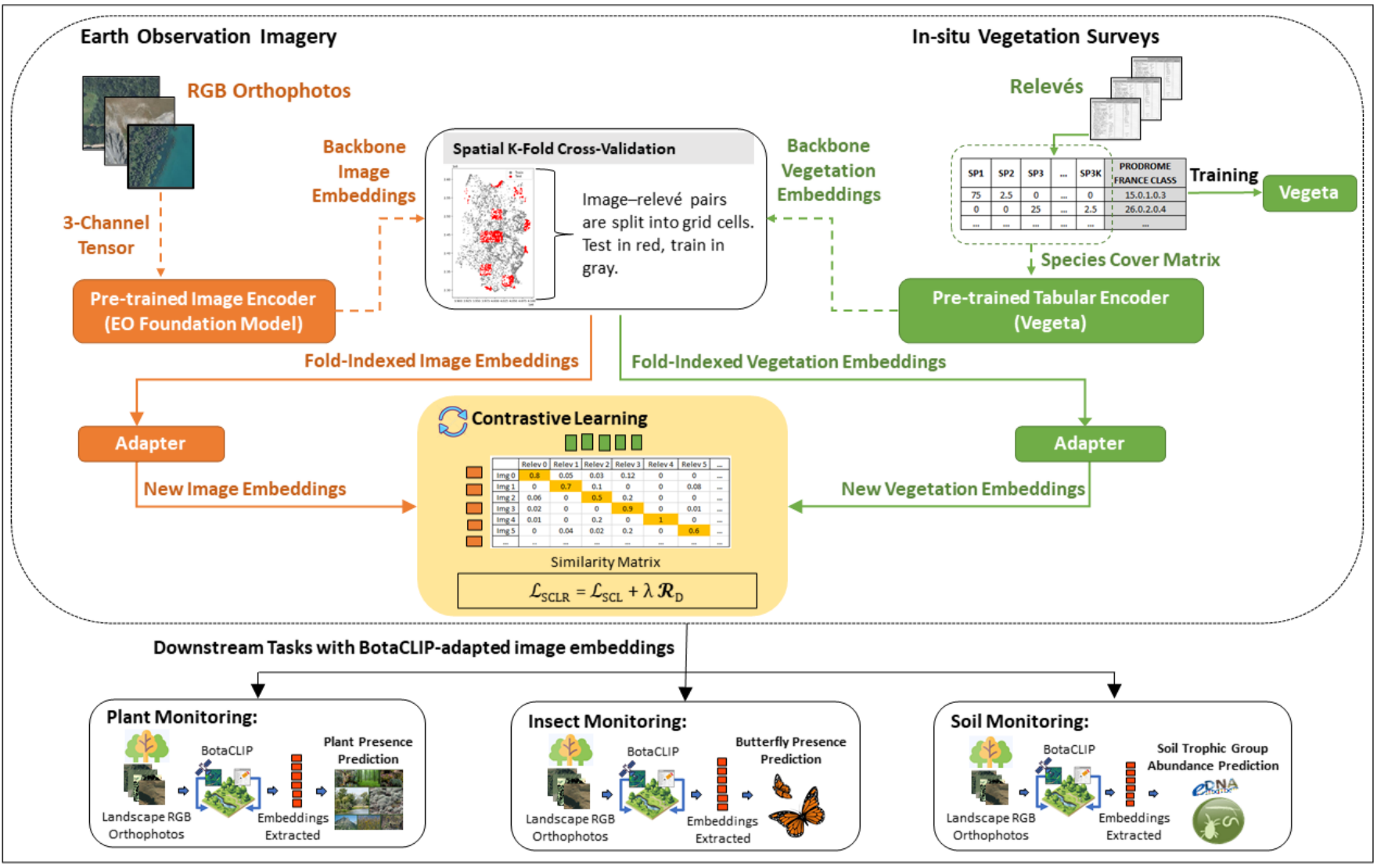}
\caption{BotaCLIP is a two modality framework: EO imagery and in-situ vegetation surveys. On the image side, RGB orthophotos are encoded into embeddings using a pretrained EO vision models (DOFA, DINOv3, or FLAIR), while in-situ vegetation relevés are encoded with a pretrained tabular encoder (VEGETA). Paired image–relevé samples are split via spatial k-fold cross-validation. During training, lightweight adapters project both modalities into a shared space and are optimized with a regularized sigmoid contrastive objective. After training, BotaCLIP image embeddings are extracted for new orthophotos and used in downstream plant, insect (butterfly), and soil prediction tasks.}
\label{MAT_botaclip_overview}
\end{figure}


BotaCLIP is a multimodal framework for botany-aware transfer. It aligns EO imagery with in-situ vegetation surveys (Figure~\ref{MAT_botaclip_overview}). Aerial RGB orthophotos are encoded into  image embeddings using a pretrained EO foundation model, while relevés represented as species–cover matrices are encoded with a lightweight tabular model. In our experiments, we use DOFA, DINOv3, and FLAIR as image encoder, and a simple MLP, hereafter referred \textit{VEGETA} for the tabular branch. We form paired image–relevé embeddings and assign them to train/test folds under a spatial k-fold cross-validation scheme. Both streams pass through linear adapters and are projected into a shared latent space, where paired samples are aligned with a sigmoid contrastive loss ($\mathcal{SCL}$). The contrastive loss is further regularized using the similarity structure of the backbone image embeddings ($\mathcal{\lambda R_D}$), encouraging the adapted space to preserve the geometry of the original EO representations while injecting vegetation-driven semantics.

The resulting space produces complementary image- and tabular-based representations. In practice, we focus on image embeddings for downstream evaluation, as aerial imagery is widely available and scalable while vegetation surveys remain costly and spatially sparse. This makes image-based embeddings the most realistic entry point for biodiversity applications at large spatial scales.

\subsection{Data Modalities}
\label{sec:data}


The BotaCLIP framework integrates two complementary modalities from the French Alps: aerial RGB orthophotos and in-situ vegetation relevés.
\paragraph{Earth Observation Imagery}
We used high-resolution aerial orthophotographs from the BD ORTHO\textsuperscript{\textregistered} dataset~\citep{bdortho} (IGN), geometrically rectified and updated every 3–4 years at 20cm resolution. For each vegetation plot (30m $\times$ 30m), we extracted a 100m$\times$100m orthophoto, yielding 28,418 RGB images. 

\paragraph{In-situ Vegetation Surveys}
The second modality comprises 28,418 relevés from the Conservatoire Botanique National Alpin~\citep{cbna}, reporting the abundance of 3,587 plant species as tabular data using the Braun-Blanquet cover-abundance scale. The Braun-Blanquet classes were converted to mean percentage values, harmonizing field estimates into continuous inputs. Each relevé was assigned to one of 232 vegetation classes in the \textit{Prodrome des Végétations de France}~\citep{prodromeVegetationsFrance}, forming a species-by-plot cover matrix (28,418 × 3,588) with an associated categorical label.

Each relevé was georeferenced, enabling pairing with its orthophoto through spatial co-registration providing aligned image–tabular samples for training.

\subsection{Embedding Extractors} 
\label{sec:backbones}


BotaCLIP relies on pretrained encoders to obtain fixed representations for both modalities prior to multimodal alignment. On the image side, we considered three pretrained Earth Observation (EO) backbones covering representative pretraining regimes: multispectral-aware EO foundation modeling (DOFA), large-scale self-supervised visual pretraining (DINOv3), and a domain-matched aerial segmentation model based on the same imagery source as in our study (FLAIR). On the tabular side, we used a local lightweight encoder pretrained to classify phytosociological vegetation classes from species-cover vectors. All backbones were kept frozen and only lightweight adapters were trained in the alignment stage.

\paragraph{DOFA~\citep{dofa}}
Dynamic One-For-All (DOFA) is a multimodal EO foundation model designed to accommodate heterogeneous sensor configurations by explicitly conditioning the visual pipeline on band/wavelength information. This design targets transfer across datasets with different spectral setups, enabling a single pretrained model to operate on varying band subsets. In our setting, we used DOFA as an RGB encoder by providing the corresponding wavelengths and extracting intermediate representations from the pretrained network. Concretely, we resized orthophotos to $224\times224$, normalized them with dataset-specific RGB statistics, and ran the checkpoint \emph{dofa\_ViT\_large\_e100.pth}. We captured embeddings using a forward hook on the \emph{head\_drop} module (i.e., just before the final prediction head), yielding one vector per image.

\paragraph{FLAIR~\citep{flair}}
French Land cover from Aerospace ImageRy (FLAIR) provides pretrained weights for aerial image understanding through semantic segmentation training, using a U-Net architecture with a ResNet34 encoder backbone. Such dense-prediction pretraining emphasizes local texture and spatial structure that can be beneficial for very-high-resolution orthophotos. For BotaCLIP, we loaded the pretrained U-Net checkpoint flair\_inc\_rgb\_15cl\_resnet34\_unet\_weights.pth and extracted embeddings from the encoder. First, we normalized RGB inputs with the preprocessing statistics associated with the released weights, forwarded the image through the network, and tapped the representation at \emph{encoder.layer4}. Since this layer outputs a spatial feature map, we applied global average pooling over spatial dimensions to obtain a single embedding vector per orthophoto.

\paragraph{DINOv3~\citep{dinov3}}
Self-Distillation with No Labels (DINOv3) provides a strong generic visual representation learned through self-supervised pretraining with Vision Transformers, yielding features that transfer well across tasks without requiring labeled EO objectives. For BotaCLIP, we use dinov3\_vitl16\_pretrain\_sat493m.pth checkpoint and extract features from late transformer blocks, which are commonly used as transferable descriptors. In our implementation, each orthophoto is resized to $224\times224$ and normalized with standard ImageNet statistics. We then extracted the outputs of the last two blocks and retained the penultimate one with layer normalization enabled, and applied mean pooling over patch tokens to obtain a single global embedding per image.

\paragraph{VEGETA}
VEGetation model for Ecosystem Typology Assignment (VEGETA) is a local MLP trained for phytosociological classification from in-situ vegetation relevés previously treated in~\ref{sec:data}. Given a $3587$-dimensional species--cover vector, VEGETA predicts the associated vegetation typology label among the $232$ classes of the \textit{Prodrome des Végétations de France}~\citep{prodromeVegetationsFrance}. To remain consistent with BotaCLIP, we trained and evaluated VEGETA using the same train/validation split defined by our spatial fold partition. VEGETA uses a feed-forward architecture of the form:
\[
3587 \xrightarrow{\text{Linear}} 1024 \xrightarrow{\text{GELU}} \xrightarrow{\text{Dropout}}
\xrightarrow{\text{Linear}} 512 \xrightarrow{\ell_2\text{-Norm}}
\xrightarrow{\text{Dropout}} \xrightarrow{\text{Linear}} 232.
\]
The model was trained with cross-entropy loss using Adam with early stopping $100$, reaching top-1$=$68\% and top-3$=$88\% accuracy on the validation split. For BotaCLIP, we extracted the $\ell_2$-normalized 512-dimensional embeddings from the last hidden layer and used it as the tabular modality in the contrastive alignment stage.

\subsection{Architecture and Contrastive Objective}
\label{sec:architecture_contrastive}

\paragraph{Image branch}
As stated in~\ref{sec:overview}, BotaCLIP does not operate on raw images, but on frozen embeddings extracted from a pretrained EO backbone described in~\ref{sec:backbones}. For each sample $i$, we denoted the image embedding as $\mathrm{Img}_i \in \mathbb{R}^{d_I}$ (e.g., $d_I=1024$ for DINOv3, $d_I=512$ for FLAIR, and $d_I=1024$ for DOFA). These embeddings were processed by a lightweight image adapter $A^{\mathrm{img}}$ with learnable parameters $\theta_{\mathrm{img}}$. In our best configuration, this adapter was implemented as a single linear layer mapping backbone features to the shared space $A^{\mathrm{img}}_{\theta_{\mathrm{img}}}: \mathbb{R}^{d_I} \rightarrow \mathbb{R}^{d},$ $d=512$. The adapted image representation was defined as $z_i^{\mathrm{img}} = \ell_2\!\big(A^{\mathrm{img}}_{\theta_{\mathrm{img}}}(\mathrm{Img}_i)\big)$, where $\ell_2(\cdot)$ denotes $\ell_2$ normalization. 


\paragraph{Vegetation branch}
On the vegetation side, species-cover vectors were first encoded by VEGETA into tabular embeddings $\mathrm{Tab}_i \in \mathbb{R}^{d_T}$, with $d_T=512$. These frozen tabular embeddings were then mapped to the shared space through a lightweight tabular adapter $A^{\mathrm{tab}}$ with learnable parameters $\theta_{\mathrm{tab}}$. In our best configuration, $A^{\mathrm{tab}}_{\theta_{\mathrm{tab}}}$ was implemented as a single linear layer
$A^{\mathrm{tab}}_{\theta_{\mathrm{tab}}}: \mathbb{R}^{d_T} \rightarrow \mathbb{R}^{d}$, with $d=512$.
The adapted tabular representation was defined as
$z_i^{\mathrm{tab}} = \ell_2\!\big(A^{\mathrm{tab}}_{\theta_{\mathrm{tab}}}(\mathrm{Tab}_i)\big)$,
where $\ell_2(\cdot)$ denotes $\ell_2$ normalization.

\paragraph{Sigmoid Contrastive Objective} 
At its core, BotaCLIP relies on the sigmoid contrastive loss~\citep{zhai2023_sigmoidloss} to align paired image–relevé samples while contrasting mismatches. The image and vegetation branches output $\ell_2$-normalized embeddings $z_i^{\mathrm{img}}, z_i^{\mathrm{tab}} \in \mathbb{R}^{d}$, so that cross-modal similarities can be measured in the shared space. Given a minibatch of $N$ paired samples, we compute all pairwise similarities between image and tabular projections. For two vectors $z, z' \in \mathbb{R}^{d}$, we denote their dot product by $z \cdot z'$. Pairwise logits are then computed as:
\begin{equation}
\ell_{ij}(\theta) = (\, z_i^{\text{img}} \cdot z_j^{\text{tab}} \,) \, \exp(\tau) + b,
\end{equation}
where $\tau$ is a learnable temperature, $b$ a learnable bias, and 
$\theta = (\theta_{\text{img}}, \theta_{\text{tab}}, \tau, b)$ collects all learnable parameters.  
We construct labels $\omega_{ij}=+1$ for positive pairs ($i=j$) and $\omega_{ij}=-1$ otherwise. Then, being $\sigma(\cdot)$ the logistic sigmoid, the sigmoid contrastive loss is:
\begin{equation}
\mathcal{L}_{\text{SCL}}(\theta) = - \frac{1}{N^2} \sum_{i=1}^N \sum_{j=1}^N 
\log \sigma \big( \omega_{ij} \, \ell_{ij}(\theta) \big),
\end{equation}

where positive pairs are pulled together and negatives are pushed apart in the shared space. However, relying solely on $\mathcal{L}_{\text{SCL}}$ can over-specialize the image space toward vegetation alignment at the cost of broader EO semantics, motivating the regularizer introduced below.

\paragraph{Geometry-preserving regularization}
Since the image embeddings $\text{Img}_i$ are extracted from a pretrained encoder, they already contain meaningful semantic structure. Our goal is to enrich them with vegetation information without discarding this prior knowledge. Relying solely on the contrastive loss $\mathcal{L}_{\text{SCL}}$ can lead to an over-specialization effect, akin to catastrophic forgetting~\citep{mccloskey1989catastrophic}. Mathematically, the optimization drives $z_i^{\text{img}}$ to match $z_i^{\text{tab}}$, reshaping the image space around dimensions that distinguish relevés while collapsing others that carry no gradient signal. Ecologically, this means that cues captured by the EO encoder but not strongly linked to vegetation composition (e.g., soil, relief, or anthropogenic patterns) risk being discarded, reducing the transferability of the embeddings to broader EO tasks.

To counterbalance this effect, we introduce a regularization term that encourages the projected embeddings $z_i^{\text{img}}$ to preserve the local similarity structure of the original EO encoder embeddings $\text{Img}_i$. Rather than enforcing $z_i^{\text{img}} \approx \text{Img}_i$ directly, we constrain pairs that were close in the EO encoder space to remain close after projection. Formally, using $\ell_2$-normalized frozen embeddings $\hat{\mathrm{Img}}_i=\mathrm{Img}_i/\|\mathrm{Img}_i\|_2$, we define the backbone and projected similarity matrices as $S_{ij}=\hat{\mathrm{Img}}_i\cdot\hat{\mathrm{Img}}_j$ and $S'_{ij}=z_i^{\mathrm{img}}\cdot z_j^{\mathrm{img}}$, respectively. We assign larger weights to pairs that are already similar in the backbone space, while avoiding duplicate pairs by retaining only the upper-triangular part and the diagonal, as in our implementation:
\begin{equation}
W_{ij} \;=\; \mathbf{1}[i \le j]\;
\Big(\mathrm{clip}_{[0,1]}\!\big(\tfrac{1 + S_{ij}}{2}\big)\Big)^2,
\label{eq:wij}
\end{equation}

where $\mathbf{1}[i\le j]$ corresponds to \texttt{triu} and $\mathrm{clip}_{[0,1]}$ to \texttt{clamp} (used for numerical stability).
The geometry-preserving regularizer penalizes changes in image-to-image similarities after projection:
\begin{equation}
\mathcal{R}_D(\theta) \;=\; \frac{1}{N^2}\sum_{i=1}^{N}\sum_{j=1}^{N}
W_{ij}\,\big(S'_{ij}-S_{ij}\big)^2.
\label{eq:reg_geom}
\end{equation}

This regularizer encourages the projected image embeddings to retain the local neighborhood structure induced by the frozen backbone, while still allowing cross-modal alignment. It is computationally lightweight, as it only relies on dot products between already available embeddings. Finally, the overall objective combines sigmoid contrastive alignment with this regularization:
\begin{equation}
\mathcal{L}_{\mathrm{SCLR}}(\theta) \;=\; \mathcal{L}_{\mathrm{SCL}}(\theta) \;+\; \lambda\,\mathcal{R}_D(\theta),
\label{eq:sclr}
\end{equation}
where $\lambda>0$ controls the regularization strength (we set $\lambda=1$ in our best configuration).

In spirit, this is similar to recent work that stabilizes representation learning by constraining similarity structure~\citep{dinov3}. In our work, the formulation operates directly on frozen backbone embeddings and preserves sample-level neighborhoods via $W_{ij}$.

\subsection{Training Strategy}
\label{sec:training_strategy}

BotaCLIP is trained with spatial cross-validation to avoid leakage due to spatial autocorrelation~\citep{Roberts2017}. The study region is partitioned into $5\text{km}\times5\text{km}$ grid cells (ETRS89/LAEA, EPSG:3035), and each relevé is assigned to its corresponding cell. Folds are defined at the cell level, with an additional one-cell buffer around each validation fold to ensure that training samples are at least 5 km away from validation samples. For efficiency, we report results on a single fold ($k=1$), which preserves spatial separation while reflecting the practical need for downstream tasks to rely on a specific trained checkpoint rather than an ensemble of models.

Optimization used AdamW with learning rate $10^{-3}$, weight decay $10^{-2}$, batch size $256$ and the regularization coefficient fixed to $\lambda=1$, training for up to 1000 epochs with early stopping (patience $=30$). The EO encoder remained frozen, and VEGETA embeddings were precomputed and kept fixed. Training therefore updates only the lightweight projection modules (the linear image and tabular projections), together with the learnable temperature and bias of the sigmoid contrastive objective. In our best configuration, with projection dimension $d=512$, this setup trains only $\sim2.4$M parameters, compared to $\sim338.16$M for DOFA, $\sim24.44$M for FLAIR, $\sim303.23$M for DINOV3, avoiding recomputation of patch-level representations and making training inexpensive in both compute and memory. Our aim is not to release another foundation model, but to provide a practical methodology for adapting existing EO encoders with ecological knowledge, making BotaCLIP lightweight and accessible. Additional ablation studies on architectural and loss variants are reported in Section~\ref{sec:results}.

\section{Experimental Setup} 
\label{sec:setup}

\subsection{Baselines}
\label{sec:baselines}

We compare BotaCLIP against two alternatives: raw frozen backbone embeddings and a supervised baseline (BotaSP). Raw embeddings serve as the simplest reference, i.e., as features without any multimodal alignment.  BotaSP is trained on top of the same frozen image embeddings to assess whether multimodal contrastive alignment yields more transferable representations than task-driven supervision. BotaSP learns a projection head followed by an MLP classifier to predict multi-label species presence/absence from frozen EO image embeddings. The projection head is a single linear layer $f$ mapping the dimensionality of the frozen backbone embedding $d_I \rightarrow d$ (with $d=512$, matching BotaCLIP’s projection dimension for a fair comparison), and its output is $\ell_2$-normalized before classification. The classifier $g$ is a two-layer MLP
\[
512 \xrightarrow{\text{Linear}} 1200 \xrightarrow{\text{GELU}} \xrightarrow{\text{Dropout}} \xrightarrow{\text{Linear}} 3587,
\]
where the $3587$ outputs correspond to the species dimensions of the vegetation relev\'es, with binary targets obtained by thresholding cover values as $\mathbf{1}[x>0]$. The model is optimized with AdamW ($lr=0.001$, $wd=0.01$, $\text{batch size}=256$) for up to 200 epochs with early stopping ($patience=10$), using the objective $\mathcal{L}=\mathcal{L}_{\text{BCE}}+\lambda\,\mathcal{R}_D$, where $\mathcal{L}_{\text{BCE}}$ is binary cross-entropy with logits and $\mathcal{R}_D$ is the geometry-preserving regularizer (with $\lambda=100$). After training, we discard the MLP classifier $g$ and keep the $512$-dimensional normalized projection $f(\cdot)$ as the image representation for downstream tasks.

\subsection{Metrics}
\label{sec:metrics}

For downstream tasks we report metrics matched to each prediction setting. For the plant and butterfly presence/absence problems, we use F1 and Sensitivity, and we additionally report two widely used ecological metrics: the \textit{True Skill Statistic} (TSS), which combines sensitivity and specificity to assess presence--absence models~\citep{Allouche2006}, and the \textit{Boyce Index} (BI) for butterflies, which evaluates how well predicted presences match observed spatial distributions beyond random expectation~\citep{ecospat}. For the soil trophic-group regression task, we report MAE and Spearman’s $\rho$ to quantify absolute error and rank agreement, respectively. In figures summarizing per-unit behavior, we further annotate results with two descriptive quantities computed across species (plants/butterflies) or trophic groups (soil): the median relative change $\tilde{\%\Delta}$ between BotaCLIP and the corresponding raw backbone features, and the fraction of units improved $u^\uparrow$.

For analyzing how alignment gains manifest in the embedding space, we frame cross-modal alignment as a \emph{bidirectional} retrieval task between image and tabular embeddings. On the evaluation split ($N{=}5515$ paired samples), each image embedding retrieves a ranked list of all tabular candidates (image$\rightarrow$tabular) and, symmetrically, each tabular embedding retrieves a ranked list of all image candidates (tabular$\rightarrow$image). We then assess whether the shared space preserves ecologically meaningful land-cover structure via \emph{habitat-aware} retrieval using seven habitat classes (Heathland, Shrubland, Grassland, Peatland, Forest, Rocky slopes, Other). A retrieved candidate is considered relevant if it shares the same habitat label as the query. We report habitat hit rates \emph{Avg HabR@1} and \emph{Avg HabR@10}, which indicate whether at least one same-habitat item appears within the top-$K$ results ($K\in{1,10}$), and \emph{Avg mAP@10}, the mean Average Precision truncated at 10 computed under habitat relevance, rewarding rankings that place same-habitat candidates higher. For visualization-driven sanity checks of habitat structure, we also report the Davies--Bouldin index (DB), where lower values indicate more compact and better separated habitat groups.

\subsection{Downstream Tasks}
\label{sec:downstream_tasks}

All baselines and BotaCLIP embeddings were evaluated on three downstream applications: (i) Plant monitoring: plant species prediction from vegetation relev\'es, (ii) Insect monitoring: butterfly species prediction from occurrence records, and (iii) Soil monitoring: soil trophic-group abundance prediction. In all cases, georeferenced biological annotations (plant cover binarized to presence/absence, butterfly presence/absence from occurrence records with pseudo-absences, or soil trophic-group relative abundances) were paired with BD ORTHO\textsuperscript{\textregistered} aerial orthophotos (20\,cm resolution, cropped to $100\times100$\,m), from which image embeddings were extracted.

Downstream models are Random Forests from Scikit-learn~\citep{pedregosa2011scikit} with default hyper-parameters. For plants, insects, and soil, experiments were repeated over 10 seeds with Stratified K-Fold cross-validation ($K=1$ for plants, $K=5$ for butterflies, and $K=5$ for soil). Results are averaged across seeds and folds. We chose this simple pipeline to match common ecological practice, which relies on libraries such as BioMod2~\citep{biomod2}. This also ensures that performance differences reflect embedding quality rather than downstream model complexity.

\paragraph{Plant Monitoring: Plant Presence Prediction}
\noindent \\
\textit{Dataset}: We used the same set of 28,418 relevés from the French Alps (3,587 species) employed to train BotaCLIP. This task is not a retraining of the model, but an explicit test of transfer: we evaluate whether image embeddings alone retain the botanical information aligned from relevés during contrastive learning. Species–cover values were binarized into presence (value $>0$) or absence ($=0$), yielding true absence information unlike pseudo-absence strategies (when we don't know if the species was actually missing or just not observed). 
To ensure sufficient support, we retained only species with at least 1,000 presences. Following the spatial split defined for BotaCLIP, we used fold $k=1$ to keep training and validation spatially disjoint. To balance classes, absences were downsampled to match presences in sets. \\
\textit{Target}: Predict binary presence/absence labels for each plant species. \\ 
\textit{Metrics}: TSS, F1, and Sensitivity.

\paragraph{Insect Monitoring: Butterfly Presence Prediction}
\noindent \\
\textit{Dataset}: Butterfly occurrence records were compiled from GBIF, restricted to human observations (2000–2022) with spatial precision $\leq$1 km, and cleaned with the CoordinateCleaner R package~\citep{Zizka2019}. We retained only records within the French Alps, discarding those below 250m elevation (urban/industrial areas) and species with fewer than 100 or more than 1,000 presences, keeping 134 species in total. The former lack statistical power, while the latter are highly generalist and ubiquitous, making their presence hard to predict from local imagery. Restricting to this intermediate range yields species with sufficient data and stronger ecological signal. Presence/absence datasets were built using pseudo-absences: occurrences marked as presences, and all other coordinates as candidate absences, downsampled to match presences for class balance. We applied a spatial 5-fold split with 5 km cells and a 1-cell buffer to avoid leakage. \\
\textit{Target}: Predict binary presence/absence labels for each butterfly species. \\
\textit{Metrics}: TSS, BI, F1, and Sensitivity.

\paragraph{Soil Monitoring: Soil Trophic Group Abundance Prediction}
\noindent \\
\textit{Dataset}: We used soil eDNA data from the French Alps long-term observatory ORCHAMP~\citep{Thuiller2024}, as detailed in~\citep{CaldernSanou2022}. 
Between 2016 and 2020, 953 soil samples were collected across 26 elevational gradients and processed with multi-marker DNA metabarcoding, yielding relative abundances for 51 trophic groups spanning biological categories: Bacteria, Fungi, Protist, Oligochaete, Insect, Collembola, Metazoa. Abundances were normalized within samples (relative proportions) and across samples (min–max scaling), and cross-validation folds were stratified using elevation quintiles to preserve the altitudinal distribution. \\
\textit{Target}: Predict continuous abundances per trophic group. \\
\textit{Metrics}: MAE and Spearman's $\rho$.

\subsection{Ablations}
\label{sec:ablations}


We ablated two core design axes of BotaCLIP while keeping the overall training pipeline fixed (frozen EO backbone, frozen VEGETA embeddings, and a shared $d{=}512$ projection space). The first axis was adapter parameterization, i.e., how image and tabular embeddings are mapped into the shared space: BMS uses nonlinear two-layer MLP adapters in both branches, BLS uses linear adapters in both branches, and BAS uses a linear image adapter coupled with an attention-based tabular adapter. The second axis was the contrastive objective: variants without the suffix R are trained with the plain sigmoid contrastive loss ($\mathcal{L}_{\mathrm{SCL}}$), whereas the R-variants (BMSR, BLSR, BASR) add our geometry-preserving regularization term ($\mathcal{L}_{\mathrm{SCLR}}$) to reduce image-space over-specialization.

This compact naming scheme makes it straightforward to interpret results: for example, BLSR denotes linear adapters trained with the regularized objective, while BAS denotes the attention-based tabular adapter trained with the plain objective. Detailed architectural definitions and optimization settings for all six variants are provided in Appendix~\ref{app:ablations}.


\section{Results}
\label{sec:results}

\subsection{Transferability and Ablations on Downstream Tasks}

\begin{figure}[!ht]
\centering
\caption{Heatmap summary of downstream transfer across EO backbones. Rows compare frozen Raw embeddings to BotaCLIP variants and columns correspond to DOFA, FLAIR, and DINOv3. Each panel reports the median performance on plants presence (TSS), butterflies occurence (BI), and soil trophic-group abundance (Spearman’s $\rho$).} 
\includegraphics[width=0.95\linewidth]{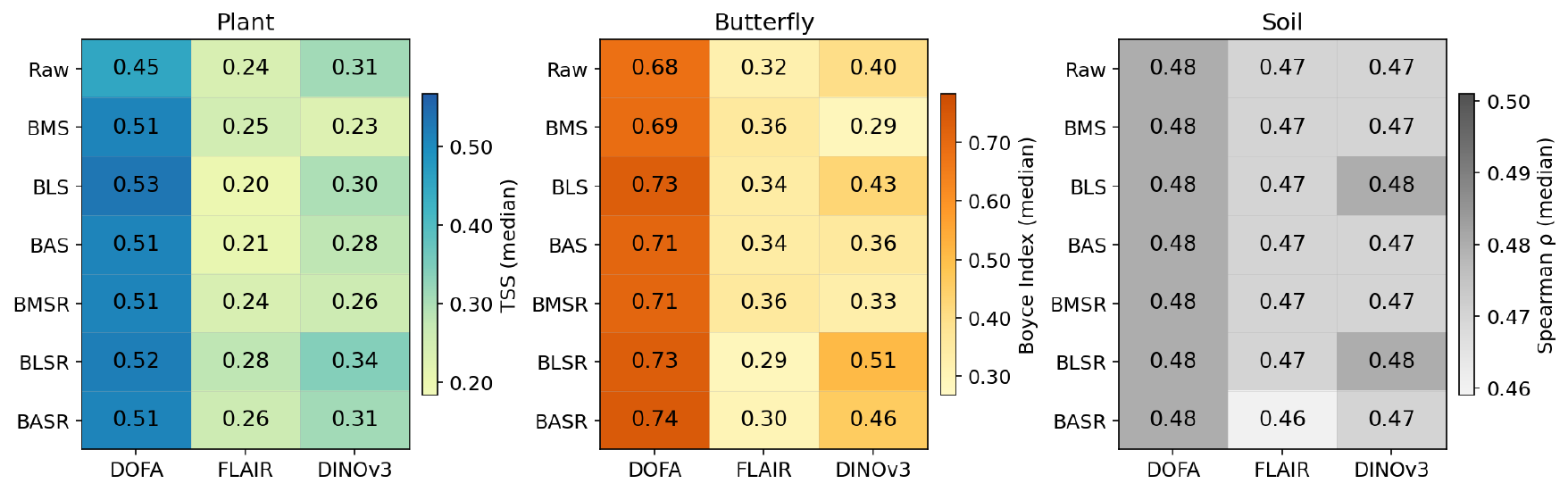}
\label{RES_downstream_tasks_summary}
\end{figure}

We first ran an embedding-space sweep to assess how sensitive alignment was to the learning rate across backbones (Appendix~\ref{app:backbone_comparison}). This diagnostic showed backbone-dependent behavior, and $lr{=}10^{-3}$ emerged as the most reliable default. Using this setting, we evaluated BotaCLIP transferability on DOFA, FLAIR, and DINOv3 across the considered architectural variants on three downstream tasks. For modeling each downstream task we used Random Forest on frozen embeddings and on BotaCLIP variants. Figure~\ref{RES_downstream_tasks_summary} reports medians aggregated over seeds and folds, computed across validation units: species for plants presence ($n{=}113$) and butterflies occurence ($n{=}134$), and trophic groups for soil ($n{=}51$). We summarize each task with one representative metric: for plants, we focus on TSS, as true presence–absence labels are available and this statistic provides a balanced evaluation of commission and omission errors under class imbalance; for butterflies, we report BI, as evaluation relies on presence–only data with pseudo-absences, making habitat suitability ranking the appropriate criterion; and for soil trophic groups, we use Spearman’s $\rho$, as the goal is to recover the relative abundance structure across functional categories rather than exact absolute values. Across backbones, DOFA yielded the strongest and most consistent gains. For plants presence, all DOFA-based BotaCLIP variants increased TSS relative to Raw (0.45), with the best median achieved by BLS (0.53). For butterflies occurrence, DOFA also benefited from alignment, with the best median Boyce Index obtained by BASR (0.74) compared to Raw (0.68). For DINOv3, improvements were more selective but still clear: BLSR yielded the best median plant TSS (0.34 vs.\ 0.31 for Raw) and the strongest gain in butterflies, reaching the highest Boyce Index (0.51 vs.\ 0.40 for Raw). FLAIR showed moderate improvements in plants (best: BLSR at 0.28 vs.\ 0.24 Raw) and mixed behavior in butterflies (best: BMS/BMSR at 0.36 vs.\ 0.32 Raw). Overall, BotaCLIP variants improved over frozen raw embeddings in most plants and butterflies settings, whereas gains for soil trophic-group abundance were consistently small (stayed in 0.46–0.48) across variants and backbones. For completeness, Appendix Tables~\ref{tab:ablation_dofa_median_iqr},~\ref{tab:ablation_flair_median_iqr}, and~\ref{tab:ablation_dinov3_median_iqr} report the full results for all metrics and variants as medians with interquartile ranges (capturing the middle 50\% of units), for each backbone.

\begin{figure}[!ht]
\centering
\caption{Performance of DOFA vs. BLSR on plants (TSS), butterflies (BI), and soil (Spearman’s $\rho$). Scatter plots (left, middle) show per-species paired scores (DOFA on $x$, BotaCLIP on $y$); points above the identity line indicate improvement. Paired dot plot with DOFA anchored at $x{=}0$ (blue) and BotaCLIP at scaled $\Delta$ (orange); shifts to the right/left indicate improved/worse trophic-group correlation. $\tilde{\%\Delta}$ and $u^\uparrow$ are the mean relative gain and the fraction of units improved, respectively, of BLSR over DOFA.}
\includegraphics[width=0.95\linewidth]{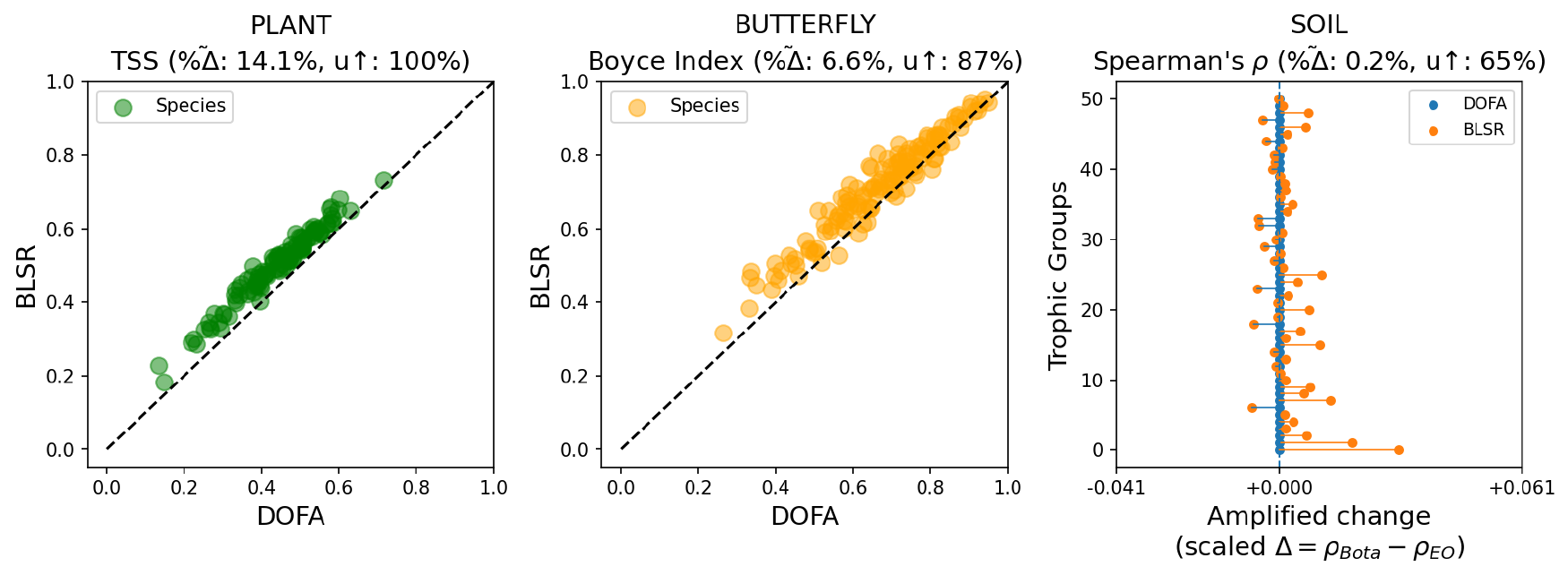}
\label{RES_downstream_tasks_Bhdx_DOFA}
\end{figure}


Since we considered several BotaCLIP variants but aimed to present a single representative configuration, we conducted a variant-selection analysis in Appendix~\ref{app:botaclip_variants_best_variant}. Briefly, within each backbone$\times$task setting we compared variants through paired, unit-level non-parametric tests and selected the configuration that offered the most consistent performance while remaining computationally efficient; this procedure identified BLSR as the representative variant. We additionally compared BLSR with the supervised baseline (BotaSP) using paired Wilcoxon signed-rank tests across validation units, and observed that BLSR outperformed BotaSP in most backbone--task settings. With BLSR fixed, we next examined whether these gains reflected broad, systematic shifts or were driven by a small subset of units by inspecting per-unit behavior in Figure~\ref{RES_downstream_tasks_Bhdx_DOFA} (additional unit-level plots for other variants are provided in Appendix~\ref{app:botaclip_variants_downstream_performances}). This unit-level breakdown also helped contextualize the small median changes observed for soil trophic groups, where performance differences were expected to be subtle across trophic groups. For BLSR (Figure~\ref{RES_downstream_tasks_Bhdx_DOFA}), most species lie above the identity line, with median unit-level relative gains of $\tilde{\%\Delta}{=}14.1\%$ (plants presence) and $6.6\%$ (butterflies occurence), and improvement rates of $u^{\uparrow}{=}100\%$ and $87\%$, respectively. For soil trophic-group abundance, changes were small ($\tilde{\%\Delta}{=}0.2\%$) with $u^{\uparrow}{=}65\%$. 


\subsection{Embedding Space Analysis Across Backbones}

\begin{figure}[!ht]
\centering

\includegraphics[width=0.99\linewidth]{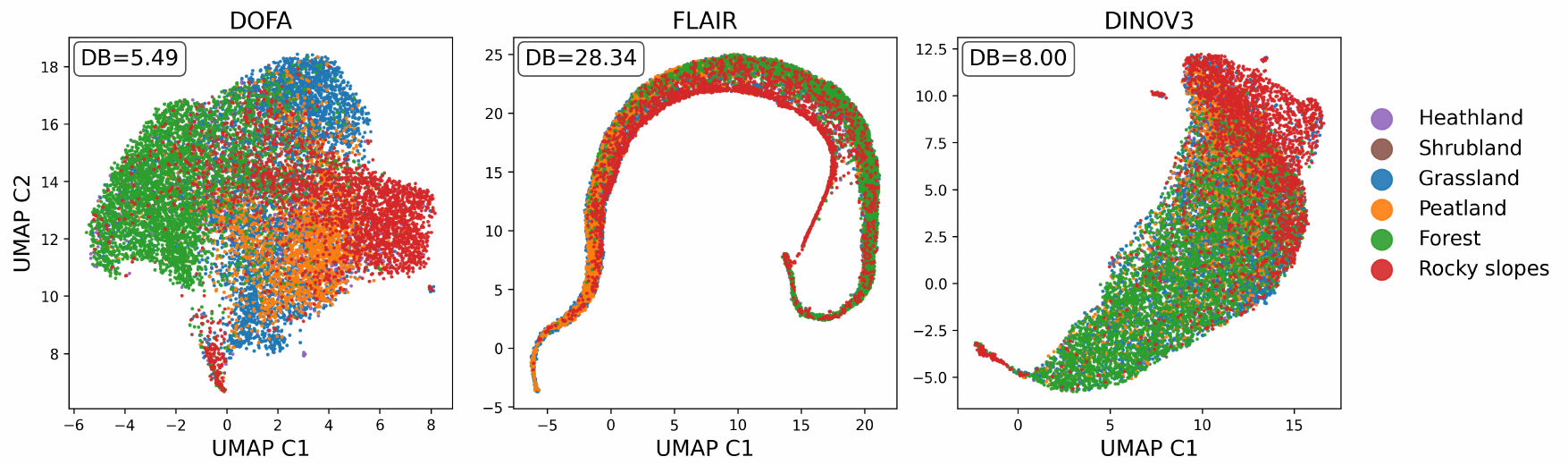}

\vspace{0.35em}

\includegraphics[width=0.99\linewidth]{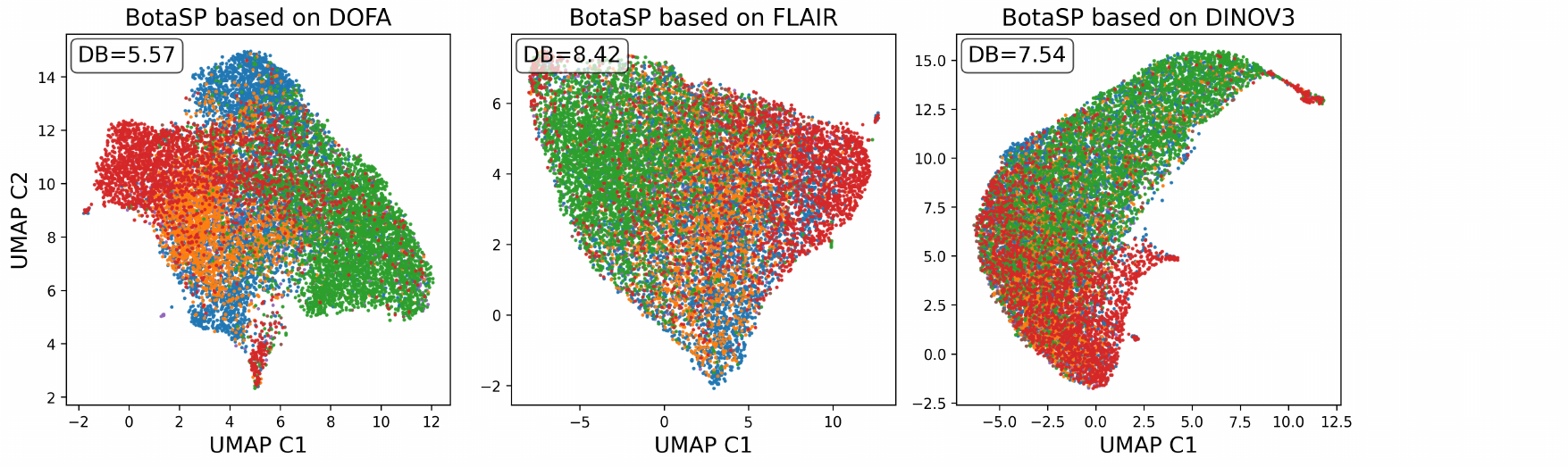}

\vspace{0.35em}

\includegraphics[width=0.99\linewidth]{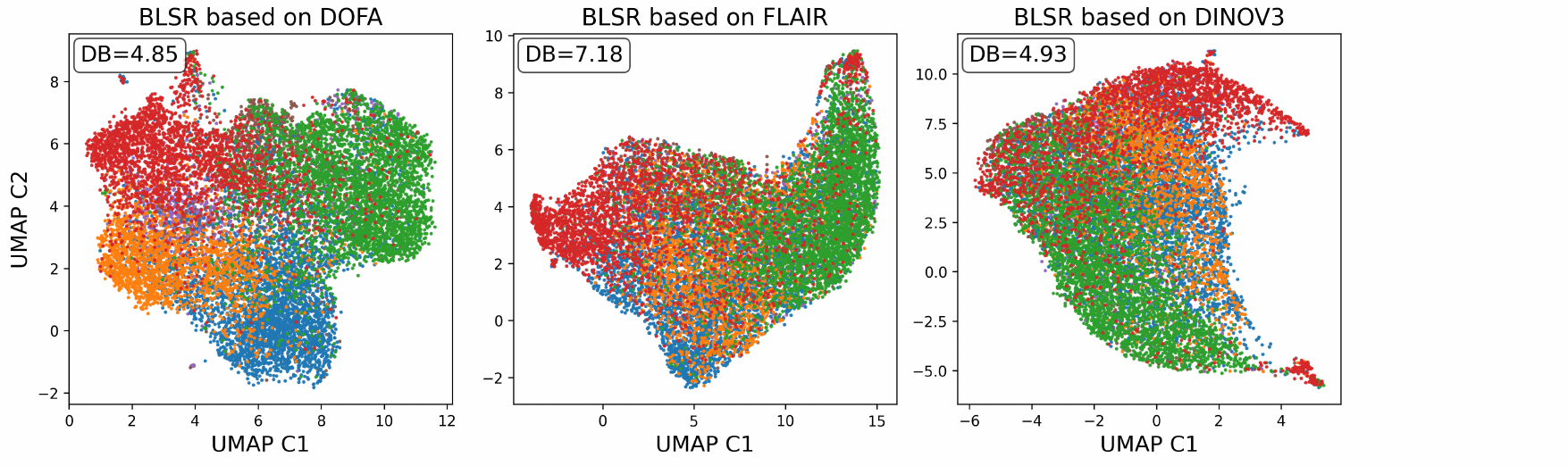}

\caption{UMAP visualization of habitat structure across EO backbones for three representations: frozen backbone embeddings (Raw, top), supervised baseline embeddings (BotaSP, middle), and BotaCLIP (BLSR, bottom). Samples from the validation set are colored by habitat class. DB index, computed on 200-dimensional PCA-reduced features, quantifies the visible structure.}
\label{embeddings_cluster_raw_botasp_blsr}
\end{figure}

To characterize how different training strategies reshape habitat-level structure in the embedding space, we visualized image embeddings in Figure~\ref{embeddings_cluster_raw_botasp_blsr} from three representations: frozen backbone features (Raw), the supervised baseline (BotaSP), and the representative contrastive configuration (BLSR). For each backbone (DOFA, FLAIR, DINOv3), we computed UMAP projections on PCA-reduced features (200 dimensions) from the validation split, and colored samples by habitat class (excluding the \textit{Other} class for readability). We also reported DB index, computed on the same 200-dimensional PCA features, as a quantitative measure of habitat separability.

Across backbones, habitat separability in the frozen feature space (Raw) varied substantially, with DOFA showing the lowest DB (5.49), DINOv3 intermediate (8.00), and FLAIR markedly higher (28.34). Supervised adaptation (BotaSP) reduced DB strongly for FLAIR (28.34$\rightarrow$8.42) and slightly for DINOv3 (8.00$\rightarrow$7.54), while leaving DOFA essentially unchanged (5.49$\rightarrow$5.57). In contrast, the representative contrastive configuration of BotaCLIP (BLSR) decreased DB for all three backbones, reaching 4.85 (DOFA), 7.18 (FLAIR), and 4.93 (DINOv3). Consistent with these scores, the UMAP projections showed more compact and better separated habitat regions under BLSR, with reduced overlap between classes compared to both Raw and BotaSP. 






\section{Discussion}
\label{sec:discussion}

\subsection{Downstream transferability} 

BotaCLIP demonstrated that lightweight multimodal alignment can inject vegetation-driven semantics into frozen pretrained EO embeddings, improving transferability for ecological tasks. Across three EO backbones (DOFA, FLAIR, and DINOv3) and BotaCLIP architectural variants (Figure~\ref{RES_downstream_tasks_summary}), aligned embeddings consistently improved performance for plant presence and butterfly occurrence prediction in most backbone settings. The DOFA unit-level analysis (Figure~\ref{RES_downstream_tasks_Bhdx_DOFA}) further showed that these gains were broadly distributed across species rather than driven by a few outliers. 

Plant presence serves as a sanity check: since BotaCLIP aligns orthophotos with vegetation relevés, recovering botanical information from images alone validates the alignment’s effectiveness. Although BotaCLIP aligns continuous cover-abundance profiles rather than binary presence labels, a simple downstream Random Forest can still decode species occurrences from the aligned embeddings, confirming that community-level semantics are recoverable from imagery alone.

Butterfly occurrence is the strongest test of independent transfer: BotaCLIP never observes butterfly records during training, yet aligned embeddings improve habitat suitability ranking under a presence-only regime. This is ecologically plausible because butterfly distributions are tightly coupled to vegetation (e.g., host plants and pollination resources), and the broad gains across 134 species suggest the learned representations capture generalizable habitat structure rather than taxon-specific cues. These results complement recent multimodal ecological foundation-model efforts~\citep{Zermatten2025_EcoWikiRS,trantas2025_bioanalyst} and underscore BotaCLIP’s ability to capture cross-taxa dependencies without explicit supervision.

Soil trophic groups are the most indirect targets. At 20\,cm resolution, RGB orthophotos provide limited insight on belowground functional structure, which is primarily driven by edaphic conditions and microclimate ~\citep{Cerna2025}; accordingly, improvements were very small, but mostly positive across trophic groups (Figure~\ref{RES_downstream_tasks_Bhdx_DOFA}). This is consistent with a physical ceiling on what optical EO can contribute to subsurface prediction.

\subsection{Transferability differences across EO backbones}

The impact of botany-aware alignment is backbone-dependent. DOFA achieves the strongest retrieval at $lr{=}10^{-3}$ (Table~\ref{RES_tab_backbone_comparison_lr}), the largest DB reductions (Figure~\ref{embeddings_cluster_raw_botasp_blsr}), and the most robust gains across tasks (Figure~\ref{RES_downstream_tasks_summary}), which is in line with its EO-centric pretraining and sensor-aware design: a representation space already organized around EO-relevant structure may be easier to specialize toward ecological semantics with lightweight linear adapters. DINOv3 improves more selectively, which is plausible given that its generic self-supervised features encode a broader visual manifold; vegetation-relevant directions exist but require stronger alignment pressure to surface. FLAIR shows moderate improvements for plants but mixed results for butterflies, where the supervised baseline performs better (Block~II, Table~\ref{tab:botaclip_selection}). Its segmentation-driven pretraining emphasizes local texture and boundary cues effective for land-cover discrimination, but these may not align with the cross-taxa semantics needed for butterfly habitat prediction. FLAIR's use of global average pooling over encoder feature maps may also compress spatial information less efficiently than transformer patch tokens, limiting what a single linear projection can recover.

\subsection{Contrastive learning versus supervised pretraining} 
 For a fixed frozen backbone, contrastive alignment with vegetation relev\'es produces embeddings that transfer more effectively than a supervised pretrained approach (BotaSP) in downstream tasks (Table~\ref{tab:botaclip_selection}). A likely reason is that contrastive learning exploits the paired structure of batches and shapes a global embedding organization through relative similarities, whereas supervised pretraining optimizes discrimination for a fixed label space and can encourage task-specific boundaries that do not necessarily generalize to other ecological endpoints. Notably, butterflies occurrence under FLAIR is an exception where task-driven supervision shows a modest paired advantage. Taken together, these results indicate that lightweight ecological injection - via vegetation composition embeddings (VEGETA) and, when used, geometry-preserving regularization - can steer generic EO representations toward ecologically meaningful spaces.

This perspective also motivates the geometry-preserving regularizer. When the projection is trained only to match the tabular branch, the image space may over-specialize toward vegetation alignment and distort backbone semantics that are not directly encoded in relev\'es, a phenomenon related to catastrophic forgetting. By constraining changes in image--image similarities, our regularizer anchors the adapted image embeddings to the local neighborhood structure of the frozen backbone while still allowing cross-modal alignment. The frequent selection of regularized variants in Block~I (Table~\ref{tab:botaclip_selection}) and the choice of BLSR as a robust and efficient configuration, supports the view that controlling specialization is beneficial when injecting domain knowledge into pretrained EO representations. 
Methodologically, our regularized is related to recent stabilization strategies for multimodal alignment that aim to preserve useful pretrained structure. For instance, \textit{Three Towers} uses a frozen pretrained encoder as an anchor during contrastive training~\citep{kossen2023three}. Similarly, \textit{Ex-MCR} studies how to extend contrastive representation spaces while maintaining compatibility with existing pretrained structure~\citep{zhang2024extending}. Our approach instead directly penalizes distortion of backbone similarity relations with a lightweight, pair-reweighted objective, and is closely related in spirit to Gram anchoring losses recently proposed to stabilize representation learning, with the difference that we operate on frozen EO embeddings and emphasize local structure via similarity-based weights~\citep{dinov3}. More broadly, the paired image--relev\'e coupling places BotaCLIP in the lineage of ecological co-ordination methods (e.g. Canonical Correspondance Analysis, Co-internia analysis), with contrastive learning providing a scalable nonlinear implementation of the same coupling principle.


\subsection{Limitations and outlook} 
BotaCLIP is evaluated on a single mountain region, and broader geographic transfer under domain shift remains to be assessed. The alignment signal depends on the quality of the relevés, which may reflect uneven sampling effort and taxonomic uncertainty. The limited gains on soil trophic groups point to a physical constraint: RGB imagery simply carries weak information about belowground processes, that community-level supervision alone cannot overcome. Incorporating complementary signals such as topography or radar imagery could partially address this. Finally, VEGETA is not ablated as a component in this study, and alternative community encoders 
or pretraining strategies may yield richer tabular representations and stronger alignment. Future work should also explore tri-modal alignment incorporating environmental covariates and extension 
to multi-sensor inputs to broaden the framework's applicability.


\section{Conclusion}

We introduced BotaCLIP, a lightweight multimodal framework that injects ecological community information into pretrained EO visual representations by aligning RGB orthophotos with in-situ vegetation relev\'es via contrastive learning. To our knowledge, this is a first step toward using community composition---rather than proxy signals such as metadata or coordinates---as guidance for adapting EO embeddings toward biodiversity-relevant semantics. BotaCLIP is backbone-agnostic, enabling controlled comparisons across EO vision encoders while keeping the alignment mechanism and tabular encoder fixed. Across three ecological tasks (plant presence, butterfly occurrence, and soil trophic-group abundance), BotaCLIP embeddings consistently outperform raw backbone features and supervised baselines, supporting parameter-efficient domain adaptation as an alternative to end-to-end retraining. Beyond biodiversity, this approach illustrates how domain-specific knowledge can adapt foundation models in data-scarce sciences.

Future work includes extending BotaCLIP to multi-sensor inputs (e.g., SAR and topography) and to tri-modal alignment by incorporating environmental covariates. This will require generalizing our contrastive objective to an img--tab--env formulation with appropriate balancing or consistency constraints, and exploring alternative vegetation community encoders beyond VEGETA. We also plan to assess robustness under domain shift by transferring to new regions and exploring uncertainty-aware geometry regularization that adapts the constraint to backbone confidence.





\section*{Declaration of competing interest}
The authors declare that they have no known competing financial interests or personal relationships that could have appeared to influence the work reported in this paper.

\section*{Funding}
This work was co-funded by the European Union (Natura Connect, No: 101060429 and OBSGESSION, No.: 101134954) and the Agence Nationale pour la Recherche through the MIAI@Grenoble Alpes (ANR-19-P3IA-0003) institute and the Office Français de la Biodiversité (OFB). Content reflects only the views of the project owners

\section*{Reproducibility statement}
All details about the architecture, loss function, and training strategy are provided in Section~\ref{sec:setup}, with further information in the appendix. Code to reproduce the experiments will be released upon publication. The vegetation, butterfly, and soil datasets are derived from existing ecological surveys and will be shared in processed form subject to licensing constraints.

\section*{Declaration of generative AI and AI-assisted technologies in the manuscript preparation process}
During the preparation of this work the author(s) used ChatGPT in order to improve the clarity and readability of the manuscript (language editing and phrasing). After using this tool/service, the author(s) reviewed and edited the content as needed and take(s) full responsibility for the content of the published article.

\bibliographystyle{elsarticle-num}
\bibliography{A9_references} 

\appendix

\section{Appendix}

\subsection{Details of Ablation Study Setups}
\label{app:ablations}

\paragraph{Compared variants.}
Across all BotaCLIP variants, we keep the same training pipeline (frozen EO backbone, frozen VEGETA embeddings, contrastive alignment in a $d{=}512$ shared space) and only modify the lightweight projection modules and the loss. We consider six models:

\begin{itemize}
    \item \textbf{BMS}: BotaCLIP with modality-specific \emph{nonlinear} adapters, where both $A^{\mathrm{img}}$ and $A^{\mathrm{tab}}$ are implemented as two-layer MLPs:
    \[
    A^{\mathrm{img}}:\quad
    d_I \xrightarrow{\text{Linear}} 2d_I \xrightarrow{\text{ReLU}} \xrightarrow{\text{Dropout}} \xrightarrow{\text{Linear}} d ,
    \]
    \[
    A^{\mathrm{tab}}:\quad
    d_T \xrightarrow{\text{Linear}} 2d_T \xrightarrow{\text{ReLU}} \xrightarrow{\text{Dropout}} \xrightarrow{\text{Linear}} d ,
    \]
     The projected features $z_i^{\mathrm{img}}$ and $z_i^{\mathrm{tab}} \in \mathbb{R}^{d}$ (with $d=512$) are $\ell_2$-normalized and the adapters are trained by minimizing the sigmoid contrastive objective $\mathcal{L}_{\mathrm{SCL}}$. Optimization is performed with AdamW (weight decay $10^{-2}$) using a cosine annealing learning-rate schedule (CosineAnnealingLR, $T_{\max}=5$) and a batch size of 256. We train for up to 1000 epochs with early stopping (patience 30) based on the validation loss. Initially, the learning rate is selected from $\{10^{-2},10^{-3},10^{-4}\}$ for each backbone to account for backbone-dependent optimization sensitivity, and we use the best-performing value (typically $10^{-3}$) for the remaining ablations and downstream experiments to ensure a consistent comparison across BotaCLIP variants.

    \item \textbf{BLS}: BotaCLIP with modality-specific \emph{linear} adapters, where both $A^{\mathrm{img}}$ and $A^{\mathrm{tab}}$ are single affine projections into the shared space:
    \[
    A^{\mathrm{img}}:\quad
    d_I \xrightarrow{\text{Linear}} d,
    \qquad
    A^{\mathrm{tab}}:\quad
    d_T \xrightarrow{\text{Linear}} d ,
    \]
    The projected representations $z_i^{\mathrm{img}}$ and $z_i^{\mathrm{tab}} \in \mathbb{R}^{d}$ (with $d=512$) are $\ell_2$-normalized and the adapters are optimized with the sigmoid contrastive objective $\mathcal{L}_{\mathrm{SCL}}$. We use the same optimization protocol as in BMS (AdamW with cosine annealing, batch size 256, early stopping 30), and fix the learning rate to $10^{-3}$ for all BLS-based experiments.

    \item \textbf{BAS}: BotaCLIP with a \emph{linear} image adapter and an \emph{attention-based} tabular adapter. The image branch uses a single affine projection and for the tabular branch, we first reduce the VEGETA embedding to a head-compatible dimension, apply self-attention (4 heads) over a single-token representation, and then project to the shared space.
    \[
    A^{\mathrm{img}}:\quad d_I \xrightarrow{\text{Linear}} d,
    \]
    \[
    A^{\mathrm{tab}}:\quad
    d_T \xrightarrow{\text{Linear}} \tilde d_T \xrightarrow{\text{Self-Attn}} \tilde d_T
    \xrightarrow{\text{LayerNorm}} \xrightarrow{\text{GELU}} \xrightarrow{\text{Dropout}} \xrightarrow{\text{Linear}} d.
    \]
    The projected features $z_i^{\mathrm{img}}$ and $z_i^{\mathrm{tab}} \in \mathbb{R}^{d}$ are $\ell_2$-normalized and the adapters are optimized with the sigmoid contrastive objective $\mathcal{L}_{\mathrm{SCL}}$. We use the same optimization protocol as in BLS (AdamW with cosine annealing, batch size 256, early stopping 30), and fix the learning rate to $10^{-3}$ for all BAS-based experiments.

    \item \textbf{BMSR}: same as BMS, but optimized with $\mathcal{L}_{\mathrm{SCLR}}$.

    \item \textbf{BLSR}: same as BLS, but optimized with $\mathcal{L}_{\mathrm{SCLR}}$.

    \item \textbf{BASR}: same as BAS, but optimized with $\mathcal{L}_{\mathrm{SCLR}}$.
\end{itemize}

\paragraph{Loss ablation: $\mathcal{L}_{\mathrm{SCL}}$ vs.\ $\mathcal{L}_{\mathrm{SCLR}}$}
We isolate the effect of the geometry-preserving regularizer by comparing variants trained with the sigmoid contrastive loss $\mathcal{L}_{\mathrm{SCL}}$ (Sec.~\ref{sec:architecture_contrastive}) against their regularized counterparts trained with $\mathcal{L}_{\mathrm{SCLR}}=\mathcal{L}_{\mathrm{SCL}}+\lambda \mathcal{R}_D$. The regularizer $\mathcal{R}_D$ encourages the projected image space to preserve neighborhood relations from the frozen backbone, mitigating over-specialization and promoting transfer to downstream tasks.

\paragraph{Projection-head ablation: adapter capacity and inductive bias}
We further ablate the adapter parameterization while keeping the rest of the framework unchanged. Specifically, we compare (i) nonlinear MLP adapters (BMS/BMSR), (ii) linear adapters (BLS/BLSR), and (iii) a hybrid design with a linear image adapter and an attention-based tabular adapter (BAS/BASR), testing whether additional capacity in the tabular branch improves cross-modal alignment and downstream transfer.

\clearpage
\subsection{Ablation Results Across Backbones}
\label{app:botaclip_variants_median_performances}

\begin{table}[!ht]
\centering
\tiny
\setlength{\tabcolsep}{2pt}
\caption{Ablation across BotaCLIP variants based on DOFA emebddings. Median $\tilde{x}$ with interquartile range [IQR].}
\label{tab:ablation_dofa_median_iqr}
\begin{tabular}{llcccccccc}
\toprule
\textbf{Dataset} & \textbf{Metric} & \textbf{DOFA} & \textbf{BotaSP} & \textbf{BMS} & \textbf{BLS} & \textbf{BAS} & \textbf{BMSR} & \textbf{BLSR} & \textbf{BASR} \\
\midrule
Plant      & TSS   & 0.45 [0.13] & 0.46 [0.12] & 0.51 [0.13] & 0.53 [0.12] & 0.51 [0.12] & 0.51 [0.13] & 0.52 [0.13] & 0.51 [0.12] \\
           & F1    & 0.24 [0.09] & 0.25 [0.09] & 0.27 [0.10] & 0.27 [0.10] & 0.27 [0.10] & 0.27 [0.09] & 0.27 [0.10] & 0.27 [0.09] \\
           & Sens. & 0.73 [0.10] & 0.73 [0.10] & 0.75 [0.10] & 0.75 [0.11] & 0.75 [0.11] & 0.74 [0.10] & 0.74 [0.12] & 0.75 [0.10] \\
\midrule
Butterfly  & TSS   & 0.26 [0.18] & 0.25 [0.17] & 0.30 [0.20] & 0.33 [0.20] & 0.32 [0.20] & 0.29 [0.20] & 0.31 [0.21] & 0.31 [0.21] \\
           & BI    & 0.68 [0.20] & 0.66 [0.21] & 0.69 [0.20] & 0.73 [0.18] & 0.71 [0.18] & 0.71 [0.18] & 0.73 [0.18] & 0.74 [0.17] \\
           & F1    & 0.67 [0.07] & 0.67 [0.06] & 0.69 [0.08] & 0.70 [0.08] & 0.70 [0.08] & 0.69 [0.08] & 0.69 [0.09] & 0.69 [0.08] \\
           & Sens. & 0.77 [0.06] & 0.76 [0.06] & 0.79 [0.09] & 0.79 [0.08] & 0.79 [0.09] & 0.78 [0.07] & 0.77 [0.08] & 0.78 [0.09] \\
\midrule
Soil       & MAE   & 0.08 [0.09] & 0.08 [0.09] & 0.08 [0.09] & 0.08 [0.09] & 0.08 [0.09] & 0.08 [0.09] & 0.08 [0.09] & 0.08 [0.09] \\
           & Spear.\ $\rho$ & 0.48 [0.28] & 0.48 [0.28] & 0.48 [0.27] & 0.48 [0.27] & 0.48 [0.28] & 0.48 [0.28] & 0.48 [0.28] & 0.48 [0.28] \\
\bottomrule
\end{tabular}
\end{table}

\begin{table}[!ht]
\centering
\tiny
\setlength{\tabcolsep}{3pt}
\renewcommand{\arraystretch}{0.95}
\caption{Ablation across BotaCLIP variants based on FLAIR emebddings. Median $\tilde{x}$ with interquartile range [IQR].}
\label{tab:ablation_flair_median_iqr}
\begin{tabular}{llcccccccc}
\toprule
\textbf{Dataset} & \textbf{Metric} & \textbf{FLAIR} & \textbf{BotaSP} & \textbf{BMS} & \textbf{BLS} & \textbf{BAS} & \textbf{BMSR} & \textbf{BLSR} & \textbf{BASR} \\
\midrule
Plant      & TSS   & 0.24 [0.13] & 0.28 [0.12] & 0.25 [0.11] & 0.20 [0.10] & 0.21 [0.11] & 0.24 [0.12] & 0.28 [0.12] & 0.26 [0.12] \\
           & F1    & 0.16 [0.08] & 0.18 [0.08] & 0.16 [0.08] & 0.15 [0.08] & 0.15 [0.08] & 0.16 [0.08] & 0.18 [0.08] & 0.17 [0.08] \\
           & Sens. & 0.62 [0.10] & 0.63 [0.08] & 0.63 [0.08] & 0.60 [0.07] & 0.61 [0.06] & 0.62 [0.08] & 0.63 [0.08] & 0.63 [0.08] \\
\midrule
Butterfly  & TSS   & 0.13 [0.09] & 0.13 [0.10] & 0.15 [0.11] & 0.14 [0.12] & 0.14 [0.12] & 0.14 [0.11] & 0.13 [0.09] & 0.13 [0.09] \\
           & BI    & 0.32 [0.23] & 0.30 [0.18] & 0.36 [0.23] & 0.34 [0.22] & 0.34 [0.22] & 0.36 [0.23] & 0.29 [0.22] & 0.30 [0.22] \\
           & F1    & 0.64 [0.03] & 0.63 [0.03] & 0.63 [0.03] & 0.63 [0.03] & 0.63 [0.03] & 0.64 [0.03] & 0.63 [0.03] & 0.63 [0.03] \\
           & Sens. & 0.77 [0.05] & 0.76 [0.05] & 0.77 [0.05] & 0.77 [0.05] & 0.76 [0.05] & 0.76 [0.05] & 0.77 [0.05] & 0.77 [0.05] \\
\midrule
Soil       & MAE   & 0.08 [0.09] & 0.08 [0.09] & 0.08 [0.09] & 0.08 [0.09] & 0.08 [0.09] & 0.08 [0.09] & 0.08 [0.09] & 0.08 [0.09] \\
           & Spear.\ $\rho$ & 0.47 [0.25] & 0.46 [0.26] & 0.47 [0.26] & 0.47 [0.26] & 0.47 [0.26] & 0.47 [0.27] & 0.47 [0.28] & 0.46 [0.27] \\
\bottomrule
\end{tabular}
\end{table}

\begin{table}[!ht]
\centering
\tiny
\setlength{\tabcolsep}{3pt}
\renewcommand{\arraystretch}{0.95}
\caption{Ablation across BotaCLIP variants based on DINOV3 emebddings. Median $\tilde{x}$ with interquartile range [IQR].}
\label{tab:ablation_dinov3_median_iqr}
\begin{tabular}{llcccccccc}
\toprule
\textbf{Dataset} & \textbf{Metric} & \textbf{DINOv3} & \textbf{BotaSP} & \textbf{BMS} & \textbf{BLS} & \textbf{BAS} & \textbf{BMSR} & \textbf{BLSR} & \textbf{BASR} \\
\midrule
Plant      & TSS   & 0.31 [0.13] & 0.32 [0.14] & 0.23 [0.12] & 0.30 [0.13] & 0.28 [0.14] & 0.26 [0.13] & 0.34 [0.13] & 0.31 [0.14] \\
           & F1    & 0.18 [0.08] & 0.19 [0.08] & 0.15 [0.07] & 0.18 [0.07] & 0.18 [0.07] & 0.17 [0.07] & 0.19 [0.08] & 0.19 [0.07] \\
           & Sens. & 0.67 [0.09] & 0.68 [0.10] & 0.63 [0.09] & 0.67 [0.11] & 0.66 [0.10] & 0.64 [0.09] & 0.69 [0.10] & 0.67 [0.11] \\
\midrule
Butterfly  & TSS   & 0.16 [0.11] & 0.17 [0.12] & 0.11 [0.08] & 0.16 [0.14] & 0.15 [0.12] & 0.14 [0.11] & 0.19 [0.14] & 0.17 [0.13] \\
           & BI    & 0.40 [0.26] & 0.46 [0.28] & 0.29 [0.25] & 0.43 [0.24] & 0.36 [0.24] & 0.33 [0.25] & 0.51 [0.28] & 0.46 [0.22] \\
           & F1    & 0.64 [0.04] & 0.64 [0.04] & 0.63 [0.03] & 0.64 [0.04] & 0.64 [0.04] & 0.64 [0.04] & 0.65 [0.05] & 0.65 [0.05] \\
           & Sens. & 0.77 [0.06] & 0.76 [0.05] & 0.78 [0.05] & 0.77 [0.06] & 0.78 [0.05] & 0.78 [0.05] & 0.77 [0.07] & 0.77 [0.05] \\
\midrule
Soil       & MAE   & 0.08 [0.09] & 0.08 [0.09] & 0.08 [0.09] & 0.08 [0.09] & 0.08 [0.09] & 0.08 [0.09] & 0.08 [0.09] & 0.08 [0.09] \\
           & Spear.\ $\rho$ & 0.47 [0.27] & 0.48 [0.27] & 0.47 [0.26] & 0.48 [0.26] & 0.47 [0.27] & 0.47 [0.27] & 0.48 [0.26] & 0.47 [0.27] \\
\bottomrule
\end{tabular}
\end{table}

\clearpage
\subsection{Computational Cost and Model Size Across BotaCLIP variants and BotaSP}
\label{app:botaclip_variants_resource_consumption}

\begin{table}[!ht]
\centering
\tiny
\setlength{\tabcolsep}{4pt}
\renewcommand{\arraystretch}{0.95}
\caption{Training cost and model size across backbones. Execution time (minutes) and total parameters for each variant (learning rate $=10^{-3}$).}
\label{tab:cost_params_backbones}
\begin{tabular}{l l rr}
\toprule
\textbf{Backbone} & \textbf{Model} & \textbf{Execution time (min)} & \textbf{Total params} \\
\midrule
\multirow{7}{*}{DOFA}
  & BotaSP & 22.79 & 5,448,387 \\
  & BMS    & 30.06 & 32,562,202 \\
  & BLS    & 11.94 & 2,361,858 \\
  & BAS    & 42.85 & 66,621,442 \\
  & BMSR   & 13.35 & 32,562,202 \\
  & BLSR   & 11.90 & 2,361,858 \\
  & BASR   & 36.54 & 66,621,442 \\
\midrule
\multirow{7}{*}{FLAIR}
  & BotaSP & 31.99 & 5,186,243 \\
  & BMS    & 21.03 & 30,464,026 \\
  & BLS    & 10.21 & 2,099,714 \\
  & BAS    & 49.50 & 66,359,298 \\
  & BMSR   & 10.10 & 30,464,026 \\
  & BLSR   & 20.99 & 2,099,714 \\
  & BASR   & 47.26 & 66,359,298 \\
\midrule
\multirow{7}{*}{DINOV3}
  & BotaSP & 21.75 & 5,448,387 \\
  & BMS    & 25.68 & 32,562,202 \\
  & BLS    & 9.89  & 2,361,858 \\
  & BAS    & 34.55 & 66,621,442 \\
  & BMSR   & 11.54 & 32,562,202 \\
  & BLSR   & 16.18 & 2,361,858 \\
  & BASR   & 30.17 & 66,621,442 \\
\bottomrule
\end{tabular}
\end{table}

\clearpage
\subsection{Representative Model Selection and Comparison to the Supervised Baseline}
\label{app:botaclip_variants_best_variant}

Because the best-performing architecture depended on the backbone and task, we selected a single representative BotaCLIP variant using paired statistical ranking across variants, as shown in Block~I of Table~\ref{tab:botaclip_selection}. We first ran a global Friedman test across BotaCLIP variants, where $\chi^2_F$ denotes the Friedman test statistic and $p_F$ its significance level. We then performed a paired Wilcoxon signed-rank test between Top-1 and Top-2, where $W$ is the Wilcoxon signed-rank statistic and $p$ the corresponding $p$-value (Holm-corrected for multiple testing). We also reported the rank-biserial effect size $r_{rb}$. Across the nine backbone$\times$task settings, BLSR and BASR most often appeared in the Top-1 and Top-2 positions produced by the paired-median ranking, with BLSR being selected most frequently. Furthermore, when BLSR was selected as Top-1 against BASR, the effect sizes were large ($r_{rb}=0.76, 0.97, 0.90$), indicating that BLSR outperformed BASR on most paired validation units. 
Finally, considering the training cost and model size reported in~\ref{app:botaclip_variants_resource_consumption}, BLSR also used substantially fewer parameters than BASR (2.1--2.4M vs.\ 66.4--66.6M) with shorter execution times. Thus, BLSR offered a good robustness-efficiency trade-off and was chosen as the representative configuration for the remainder of the analysis.

In Block~II of Table~\ref{tab:botaclip_selection}, we compared the representative contrastive variant (BLSR) against the supervised baseline (BotaSP) using a paired Wilcoxon signed-rank test. Here we report the uncorrected $p$-values, i.e., without applying an additional multiple-comparisons adjustment, because each row corresponds to a single, pre-specified comparison (BLSR vs.\ BotaSP) rather than a family of post-hoc tests across many variants. In this validation, BLSR significantly outperformed BotaSP for plants across all backbones, and for butterflies under DOFA and DINOv3. In contrast, for butterflies with FLAIR the supervised baseline was better ($p{=}1.09{\times}10^{-3}$), although the associated effect size was modest ($r_{rb}{=}0.33$), indicating a consistent but not large paired advantage. For soil trophic groups, differences were small and not consistently significant, with non-significant contrasts under FLAIR and DINOv3.

\begin{table}[!ht]
\centering
\tiny
\setlength{\tabcolsep}{3.6pt}
\renewcommand{\arraystretch}{0.98} 
\caption{BotaCLIP variant selection and comparison to the supervised baseline.
Block I (intra-BotaCLIP): For each backbone$\times$task, variants are ranked by paired median performance across validation units (Top-1/Top-2); we report a Friedman test across variants ($\chi^2_F$, $p_F$), followed by a paired Wilcoxon signed-rank test comparing Top-1 vs.\ Top-2 (Holm-corrected $p$, Wilcoxon test $W$, and rank-biserial $r_{rb}$). 
Block II (contrastive vs.\ supervised): The selected representative variant (BLSR) is compared to the supervised baseline (BotaSP) with a paired Wilcoxon test over the same units (uncorrected $p$, single pre-specified comparison).}
\label{tab:botaclip_selection}
\begin{tabular}{llccccccc}
\toprule
\textbf{Backbone} & \textbf{Dataset} &
\textbf{Top-1} & \textbf{Top-2} &
$\boldsymbol{\chi^2_F}$ & $\mathbf{p_F}$ &
$W$ & $\mathbf{p}$ & $r_{rb}$ \\
\midrule

\multicolumn{9}{l}{\textbf{Block I --- Intra-BotaCLIP model selection}} \\
\midrule

\multirow{3}{*}{DOFA}
& Plant (TSS)           
& BLS  & \textbf{BLSR}
& 377.48 & $1.94{\times}10^{-78}$
& 471.00  & $1.01{\times}10^{-14}$ & 0.85 \\

& butterflies (BI)        
& BASR & \textbf{BLSR}
& 282.48 & $4.63{\times}10^{-58}$
& 4224.00 & $>0.05$ (tie) & -0.07 \\

& soil (Spear.\ $\rho$) 
& BASR & BMS
& 24.80  & $3.72{\times}10^{-4}$
& 620.00  & $>0.05$ (tie) & -0.06 \\
\midrule

\multirow{3}{*}{FLAIR}
& Plant (TSS)           
& \textbf{BLSR} & BASR
& 452.10 & $1.73{\times}10^{-94}$
& 780.00  & $5.43{\times}10^{-12}$ & 0.76 \\

& butterflies (BI)        
& BMSR & BMS
& 212.57 & $3.99{\times}10^{-43}$
& 4307.00 & $>0.05$ (tie) & 0.05 \\

& soil (Spear.\ $\rho$) 
& BMSR & BMS
& 8.64   & $1.95{\times}10^{-1}$
& 469.00  & $>0.05$ (tie) & 0.29 \\
\midrule

\multirow{3}{*}{DINOv3}
& Plant (TSS)           
& \textbf{BLSR} & BASR
& 593.31 & $6.48{\times}10^{-125}$
& 83.00   & $5.01{\times}10^{-19}$ & 0.97 \\

& butterflies (BI)        
& \textbf{BLSR} & BASR
& 511.52 & $2.77{\times}10^{-107}$
& 432.00  & $2.09{\times}10^{-19}$ & 0.90 \\

& soil (Spear.\ $\rho$) 
& \textbf{BLSR} & BLS
& 14.52  & $2.43{\times}10^{-2}$
& 583.00  & $>0.05$ (tie) & 0.12 \\
\midrule

\multicolumn{9}{l}{\hspace{1.5cm}\textbf{$\rightarrow$ Representative model (robustness + efficiency): \; \textbf{BLSR}}} \\
\midrule

\multicolumn{9}{l}{\textbf{Block II --- Representative model vs.\ BotaSP (contrastive vs.\ supervised)}} \\
\midrule

\multirow{3}{*}{DOFA}
& Plant (TSS)           
& \textbf{BLSR} & BotaSP
& -- & -- & 1.00 & $2.88{\times}10^{-20}$ & 1.00 \\

& butterflies (BI)        
& \textbf{BLSR} & BotaSP
& -- & -- & 80.00 & $5.85{\times}10^{-23}$ & 0.98 \\

& soil (Spear.\ $\rho$) 
& \textbf{BLSR} & BotaSP
& -- & -- & 439.00 & $3.58{\times}10^{-2}$ & 0.34 \\
\midrule

\multirow{3}{*}{FLAIR}
& Plant (TSS)           
& \textbf{BLSR} & BotaSP
& -- & -- & 930.00 & $5.31{\times}10^{-11}$ & 0.71 \\

& butterflies (BI)        
& \textbf{BotaSP} & BLSR
& -- & -- & 3052.00 & $1.09{\times}10^{-3}$ & 0.33 \\

& soil (Spear.\ $\rho$) 
& \textbf{BLSR} & BotaSP
& -- & -- & 656.00 & $9.48{\times}10^{-1}$ (n.s.) & 0.01 \\
\midrule

\multirow{3}{*}{DINOv3}
& Plant (TSS)           
& \textbf{BLSR} & BotaSP
& -- & -- & 725.50 & $8.82{\times}10^{-13}$ & 0.77 \\

& butterflies (BI)        
& \textbf{BLSR} & BotaSP
& -- & -- & 689.00 & $1.69{\times}10^{-17}$ & 0.85 \\

& soil (Spear.\ $\rho$) 
& \textbf{BLSR} & BotaSP
& -- & -- & 532.00 & $2.19{\times}10^{-1}$ (n.s.) & 0.20 \\
\bottomrule
\end{tabular}
\end{table}

\clearpage
\subsection{Unit-level Gains Across BotaCLIP Variants Based on DOFA}
\label{app:botaclip_variants_downstream_performances}

\begin{figure}[!ht]
\centering
\caption{Performance of DOFA vs. BotaSP on plants (TSS), butterflies (BI), and soil (Spearman’s $\rho$). Scatter plots (left, middle) show per-species paired scores (DOFA on $x$, BotaCLIP on $y$); points above the identity line indicate improvement. Paired dot plot with DOFA anchored at $x{=}0$ (blue) and BotaCLIP at scaled $\Delta$ (orange); shifts to the right/left indicate improved/worse trophic-group correlation. $\tilde{\%\Delta}$ and $u^\uparrow$ are the mean relative gain and the fraction of units improved, respectively, of BotaSP over DOFA.} 
\includegraphics[width=0.99\linewidth]{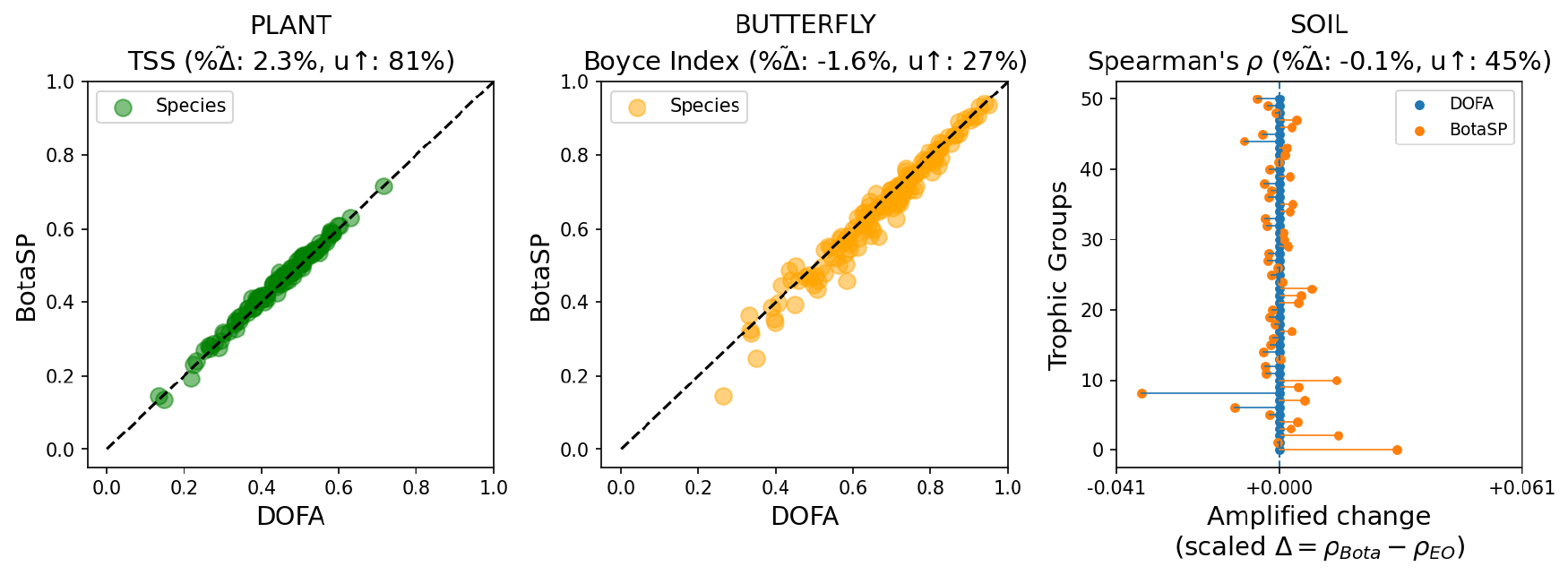}
\label{RES_downstream_tasks_Bsp_DOFA}
\end{figure}

\begin{figure}[!ht]
\centering
\caption{Performance of DOFA vs. BMS on plants (TSS), butterflies (BI), and soil (Spearman’s $\rho$). Scatter plots (left, middle) show per-species paired scores (DOFA on $x$, BotaCLIP on $y$); points above the identity line indicate improvement. Paired dot plot with DOFA anchored at $x{=}0$ (blue) and BotaCLIP at scaled $\Delta$ (orange); shifts to the right/left indicate improved/worse trophic-group correlation. $\tilde{\%\Delta}$ and $u^\uparrow$ are the mean relative gain and the fraction of units improved, respectively, of BMS over DOFA.} 
\includegraphics[width=0.99\linewidth]{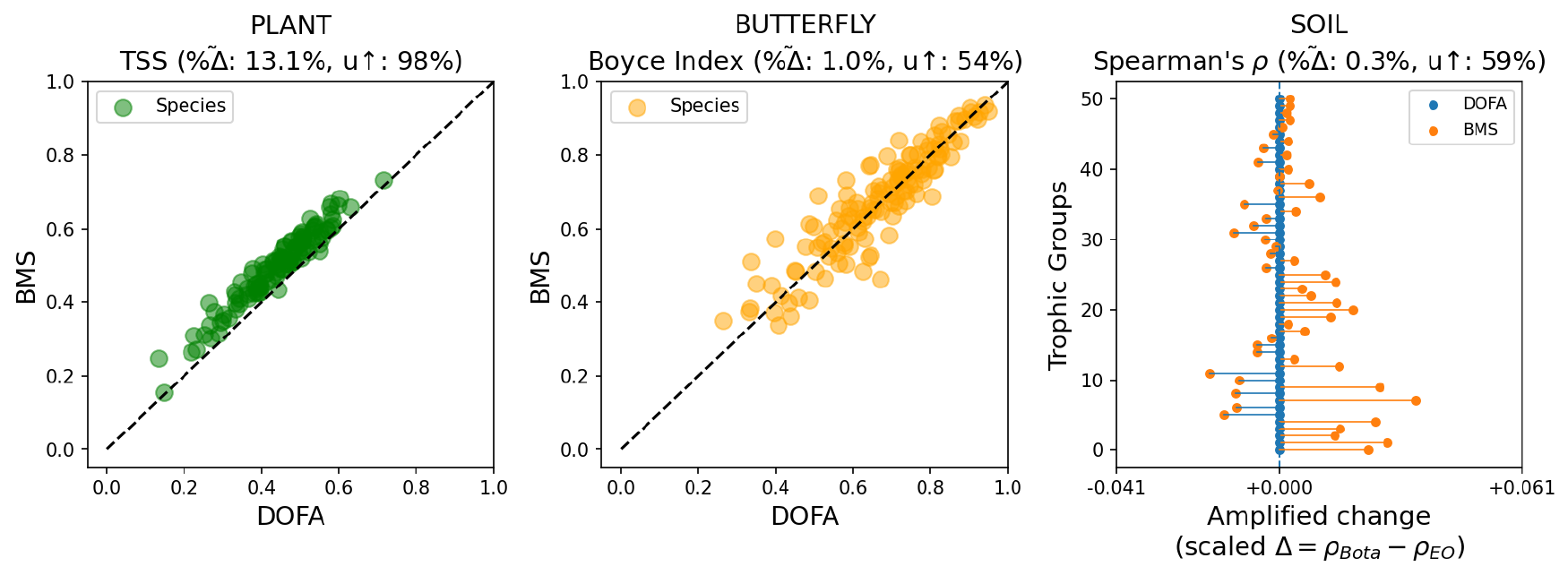}
\label{RES_downstream_tasks_Bbase_DOFA}
\end{figure}

\begin{figure}[!ht]
\centering
\caption{Performance of DOFA vs. BLS on plants (TSS), butterflies (BI), and soil (Spearman’s $\rho$). Scatter plots (left, middle) show per-species paired scores (DOFA on $x$, BotaCLIP on $y$); points above the identity line indicate improvement. Paired dot plot with DOFA anchored at $x{=}0$ (blue) and BotaCLIP at scaled $\Delta$ (orange); shifts to the right/left indicate improved/worse trophic-group correlation. $\tilde{\%\Delta}$ and $u^\uparrow$ are the mean relative gain and the fraction of units improved, respectively, of BLS over DOFA.} 
\includegraphics[width=0.99\linewidth]{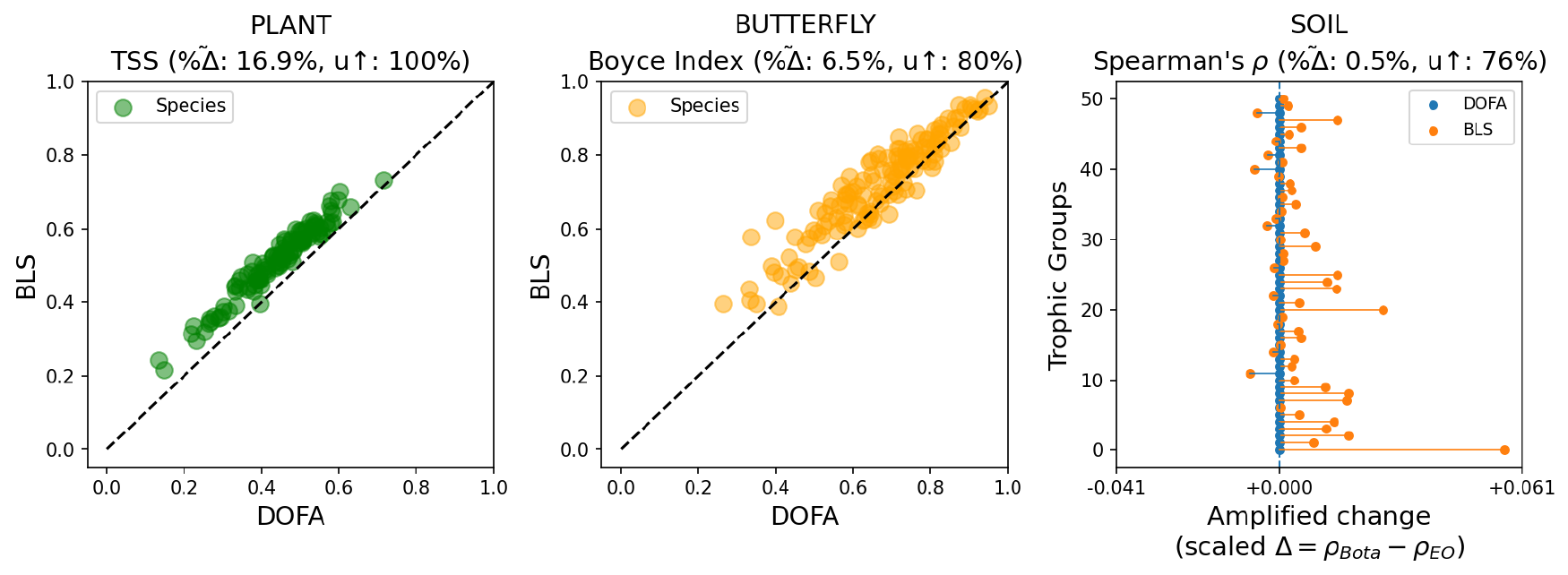}
\label{RES_downstream_tasks_Bsregulin_DOFA}
\end{figure}

\begin{figure}[!ht]
\centering
\caption{Performance of DOFA vs. BAS on plants (TSS), butterflies (BI), and soil (Spearman’s $\rho$). Scatter plots (left, middle) show per-species paired scores (DOFA on $x$, BotaCLIP on $y$); points above the identity line indicate improvement. Paired dot plot with DOFA anchored at $x{=}0$ (blue) and BotaCLIP at scaled $\Delta$ (orange); shifts to the right/left indicate improved/worse trophic-group correlation. $\tilde{\%\Delta}$ and $u^\uparrow$ are the mean relative gain and the fraction of units improved, respectively, of BAS over DOFA.} 
\includegraphics[width=0.99\linewidth]{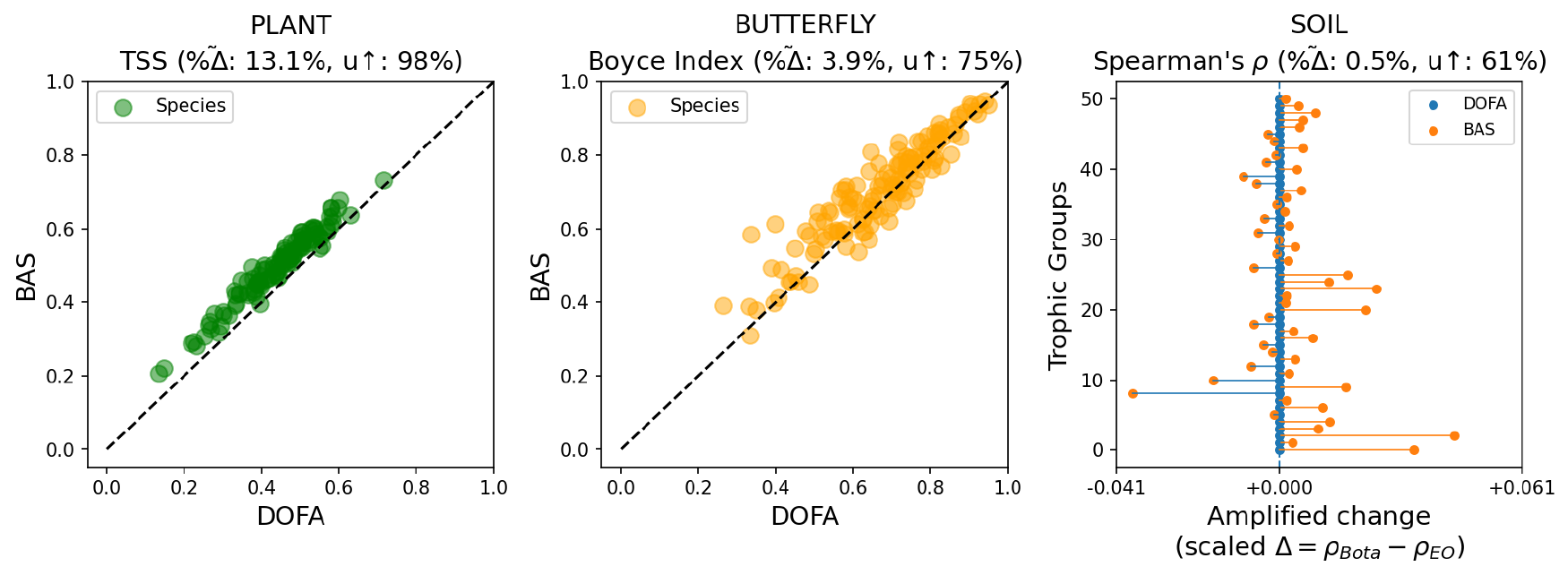}
\label{RES_downstream_tasks_Bsreguatte_DOFA}
\end{figure}

\begin{figure}[!ht]
\centering
\caption{Performance of DOFA vs. BMSR on plants (TSS), butterflies (BI), and soil (Spearman’s $\rho$). Scatter plots (left, middle) show per-species paired scores (DOFA on $x$, BotaCLIP on $y$); points above the identity line indicate improvement. Paired dot plot with DOFA anchored at $x{=}0$ (blue) and BotaCLIP at scaled $\Delta$ (orange); shifts to the right/left indicate improved/worse trophic-group correlation. $\tilde{\%\Delta}$ and $u^\uparrow$ are the mean relative gain and the fraction of units improved, respectively, of BMSR over DOFA.} 
\includegraphics[width=0.99\linewidth]{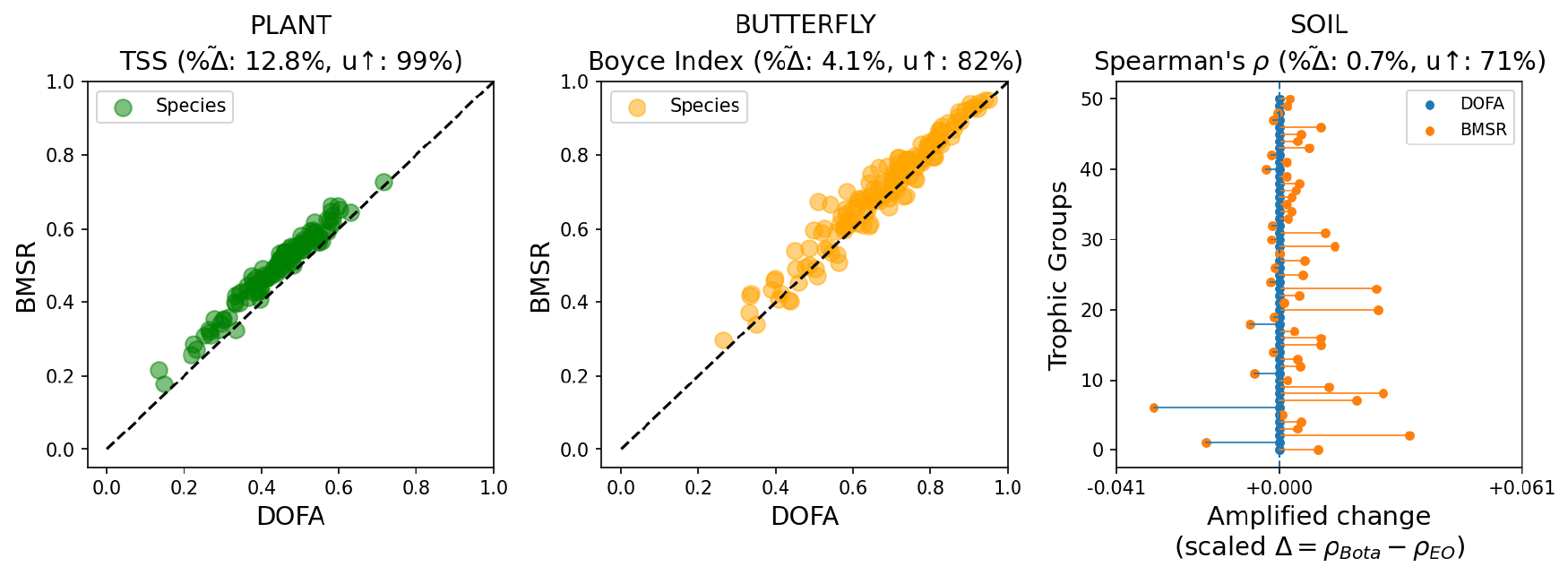}
\label{RES_downstream_tasks_Bregu_DOFA}
\end{figure}

\begin{figure}[!ht]
\centering
\caption{Performance of DOFA vs. BASR on plants (TSS), butterflies (BI), and soil (Spearman’s $\rho$). Scatter plots (left, middle) show per-species paired scores (DOFA on $x$, BotaCLIP on $y$); points above the identity line indicate improvement. Paired dot plot with DOFA anchored at $x{=}0$ (blue) and BotaCLIP at scaled $\Delta$ (orange); shifts to the right/left indicate improved/worse trophic-group correlation. $\tilde{\Delta}$ and $u^\uparrow$ are the mean relative gain and the fraction of units improved, respectively, of BASR over DOFA.} 
\includegraphics[width=0.95\linewidth]{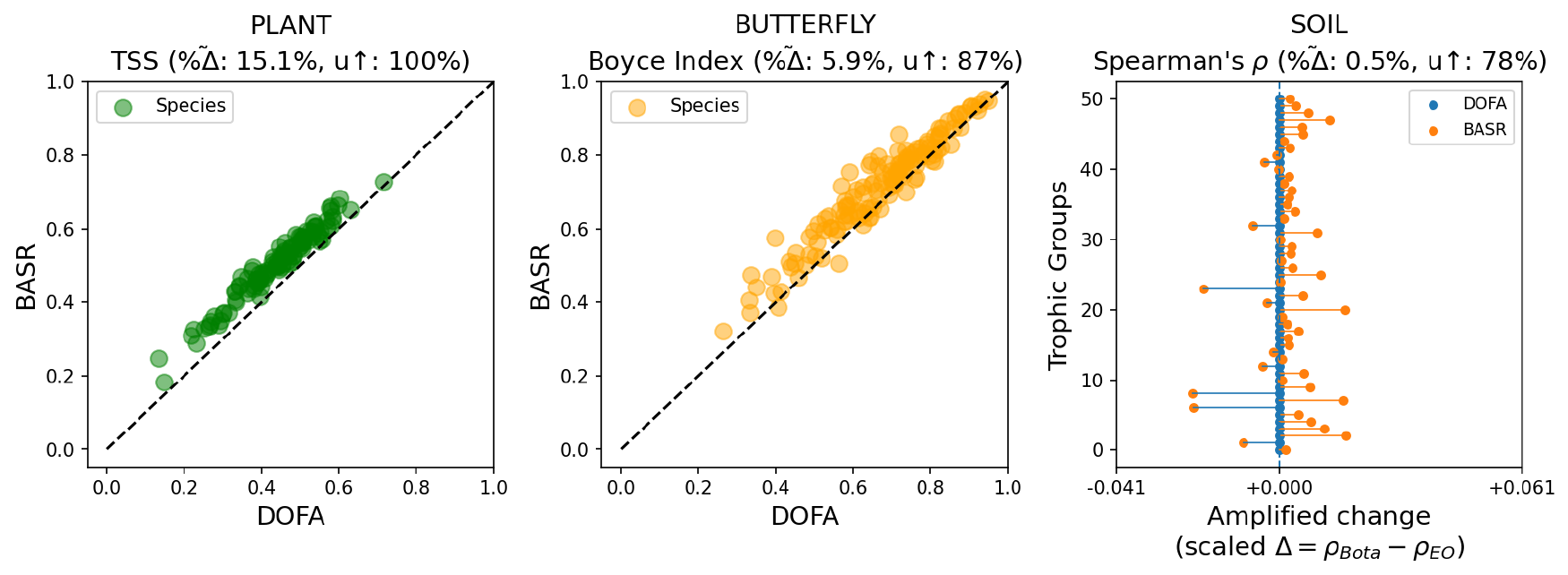}
\label{RES_downstream_tasks_Batte_DOFA}
\end{figure}

\clearpage
\subsection{Embedding-space Diagnostics Across Backbones}
\label{app:backbone_comparison}

We compared pretrained EO vision models and assessed sensitivity to the learning rate used to train BotaCLIP’s lightweight adapters. For this backbone/LR analysis, we report retrieval results with BMS, our reference configuration with nonlinear (MLP) adapters in both branches. Keeping the alignment setup unchanged (architecture and loss), so that observed differences mainly reflect the EO vision model and learning rate. Table~\ref{RES_tab_backbone_comparison_lr} reports bidirectional habitat-aware cross-modal retrieval on the validation set ($N{=}5515$), averaged over image$\leftrightarrow$tabular directions (Avg HabR@1, Avg HabR@10, Avg mAP@10). We observed two patterns. First, performance was learning-rate dependent, with the best value varying across EO vision models: DOFA peaked at $lr{=}10^{-3}$ (Avg HabR@10$=91.77$, Avg mAP@10$=60.48$), while FLAIR and DINOv3 showed smaller gains and less pronounced optima. 

\begin{figure}[!ht]
\centering
\caption{UMAP visualization of habitat structure in frozen EO backbone embeddings (top row) versus BotaCLIP-aligned image embeddings trained with BMS (bottom row) for each backbone. Samples from the validation set are colored by habitat class. DB index, computed on 200-dimensional PCA-reduced features, quantifies the visible structure.}
\includegraphics[width=0.9\linewidth]{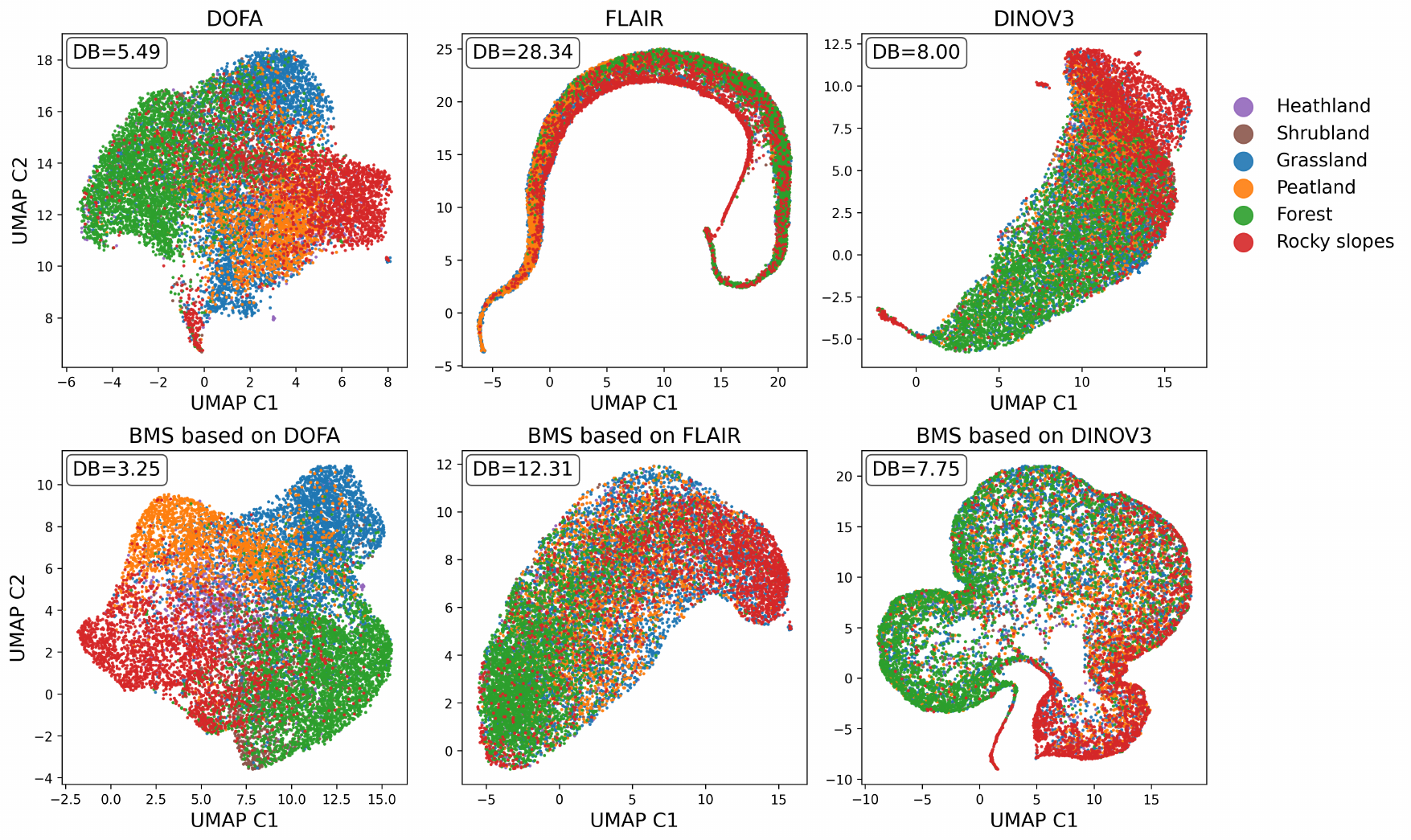}
\label{RES_embeddings_cluster_backbones_bms}
\end{figure}

Second, habitat-aware retrieval scores are consistently high across settings, suggesting that alignment primarily improves land-cover organization in the joint space. This behavior was also visible in the UMAP sanity check in Figure~\ref{RES_embeddings_cluster_backbones_bms}, which visualizes habitat organization and reports DB index as a quantitative proxy of habitat separability. For readability, the UMAP plots exclude the \textit{Other} habitat class, whereas retrieval metrics in Table~\ref{RES_tab_backbone_comparison_lr} are computed on the full pool including \textit{Other}. DB scores are computed on PCA-reduced features (200 dimensions) rather than on the 2D UMAP projection. Across the three backbones, BotaCLIP (BMS) reduced DB relative to the frozen features, with the clearest gain for DOFA (5.49 $\rightarrow$ 3.25) and FLAIR (28.34 $\rightarrow$ 12.31), while DINOv3 showed a smaller but consistent decrease (8.00 $\rightarrow$ 7.75).

Based on these results, we use $lr{=}10^{-3}$ as the default choice in subsequent experiments, while still reporting results across all three backbones to evaluate the backbone-agnostic behavior of BotaCLIP.

\begin{table}[!ht]
\centering
\tiny
\caption{Habitat-aware cross-modal retrieval for BotaCLIP (BMS variant) averaged over image$\leftrightarrow$tabular directions on the validation set, across EO vision models and learning rates. Relevance is defined by 7 habitat classes; we report habitat hit rates (Avg HabR@1, Avg HabR@10) and habitat-aware ranking quality (Avg mAP@10).}
\setlength{\tabcolsep}{3.2pt}
\begin{tabular}{llccc}
\toprule
\multirow{2}{*}{Backbone} & \multirow{2}{*}{LR} &
\multicolumn{3}{c}{Habitat-aware (\%)} \\
\cmidrule(lr){3-5}
& & Avg HabR@1 & Avg HabR@10 & Avg mAP@10 \\
\midrule
\multirow{3}{*}{DOFA}
& $10^{-2}$ & 32.72 & 68.28 & 42.24 \\
& $10^{-3}$ & \textbf{51.58} & \textbf{91.77} & \textbf{60.48} \\
& $10^{-4}$ & 51.39 & 79.29 & 44.75 \\
\midrule
\multirow{3}{*}{FLAIR}
& $10^{-2}$ & 39.89 & 72.41 & 44.07 \\
& $10^{-3}$ & \textbf{43.45} & \textbf{82.02} & \textbf{48.43} \\
& $10^{-4}$ & 42.60 & 75.24 & 45.54 \\
\midrule
\multirow{3}{*}{DINOv3}
& $10^{-2}$ & \textbf{42.67} & 79.80 & 42.79 \\
& $10^{-3}$ & 33.77 & \textbf{84.50} & \textbf{44.79} \\
& $10^{-4}$ & 40.42 & 79.51 & 43.59 \\
\bottomrule
\end{tabular}
\label{RES_tab_backbone_comparison_lr}
\end{table}

\end{document}